\documentclass{article}

\PassOptionsToPackage{numbers, sort&compress}{natbib}
\usepackage[preprint]{neurips_2026}

\usepackage[utf8]{inputenc} 
\usepackage[T1]{fontenc}    
\usepackage{hyperref}       
\usepackage{url}            
\usepackage{booktabs}       
\usepackage{amsfonts}       
\usepackage{nicefrac}       
\usepackage{microtype}      
\usepackage{xcolor}         
\usepackage{graphicx}
\usepackage{amsmath}
\usepackage{caption}
\usepackage{pifont}
\newcommand{\cmark}{\ding{51}}   

\usepackage{subcaption}    
\usepackage{tabularx}
\newcolumntype{Y}{>{\centering\arraybackslash}X}
\usepackage{multirow}
\usepackage{wrapfig}
\usepackage{verbatim}
\usepackage{algorithm}      
\usepackage{setspace}       
\usepackage{amsmath}        
\usepackage{amssymb}        
\usepackage{placeins}
\usepackage{algpseudocode}
\usepackage{hyperref}
\usepackage{xcolor}

\definecolor{linkblue}{RGB}{43, 110, 230}
\hypersetup{
    colorlinks=true,
    urlcolor=black,
    linkcolor=linkblue,
    citecolor=linkblue
}

\makeatletter
\renewcommand{\@bottomtitlebar}{
  \vskip 0.29in
  \vskip -\parskip
  \hrule height 1\p@
  \vskip 0.05in
}
\makeatother

\makeatletter
\renewcommand\footnoterule{%
  \vfill
  \kern-3\p@
  \hrule\@width.4\columnwidth
  \kern2.6\p@}
\makeatother
\title{3DPhysVideo: Consistency-Guided Flow SDE for Video Generation via 3D Scene Reconstruction and Physical Simulation}
\author{%
  Hwidong Kim\thanks{Equal contribution.} \quad
  Yunho Kim\footnotemark[1] \quad
  Tae-Kyun Kim \\[1pt]
  KAIST \\[1pt]
  \texttt{\{hwidong, youknowyunho, kimtaekyun\}@kaist.ac.kr} \\[5pt]
  {\small Project page: {\hypersetup{urlcolor=linkblue}\url{https://hwidong-kim.github.io/projects/3DPhysVideo}}}
}

\begin{document}
\maketitle
\vspace{-1cm}

\begin{center}
\centering
\begin{figure}[h]
    \centering
    \includegraphics[width=1.0\textwidth]{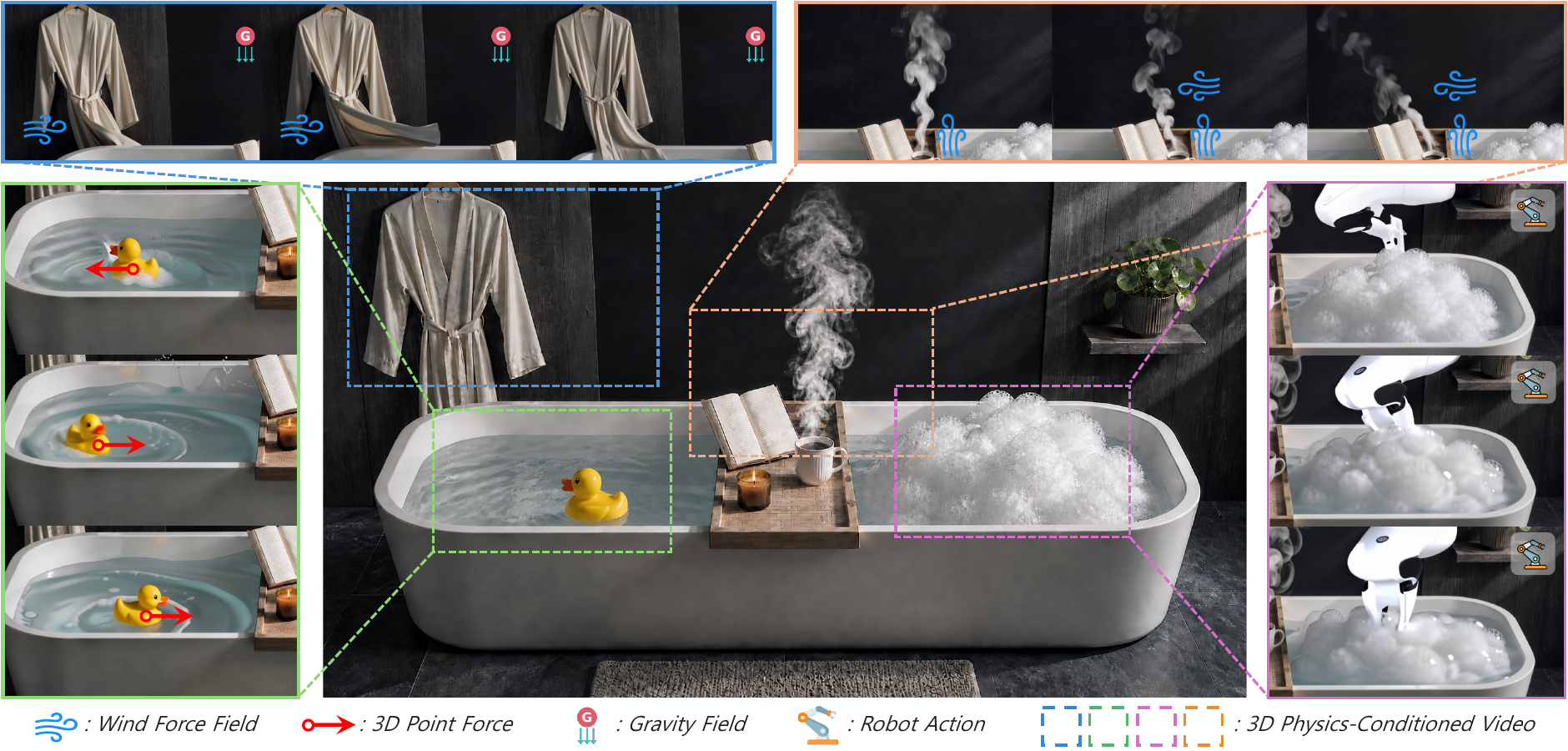}
    \caption{We propose \textsc{3DPhysVideo}, a training-free framework for 3D physics-conditioned video generation, leveraging an off-the-shelf video model. From an input scene (center), our method enables users to apply diverse physical controls to a variety of materials. See Sec.~\ref{app:teaser_details} of the supplementary for details on the simulation setup.}
    \vspace{-1.4cm}
    \label{fig:teaser}
\end{figure}
\end{center}

\begin{abstract}
\vspace{-0.7em}
Video generative models have made remarkable progress, yet they often yield visual artifacts that violate grounding in physical dynamics. Recent works such as PhysGen3D tackle single image-to-3D physics through mesh reconstruction and Physically-Based Rendering, but challenges remain in modeling fluid dynamics, multi-object interactions and photorealism. This work introduces \textsc{3DPhysVideo}, a novel training-free pipeline that generates physically realistic videos from a single image. We repurpose an off-the-shelf video model for two stages. First, we use it as a novel view synthesizer to reconstruct complete 360-degree 3D scene geometry by guiding the image-to-video (I2V) flow model with rendered point clouds. 
Second, after applying physics solvers to this geometry, the physically simulated point cloud is used to guide the same I2V flow model to synthesize final, high-quality videos. Consistency-Guided Flow SDE, which decomposes the predicted velocity of the I2V flow model into denoising and consistency bias, enforces consistency to the conditional inputs, allowing us to effectively repurpose the model for both 3D reconstruction and simulation-guided video generation. 
In the diverse experiments including multi-objects, and fluid interaction scenes, 
our method successfully bridges the gap from single-images to physically plausible videos, while remaining efficient to run on a single consumer gpu. It outperforms state-of-the-art baselines on GPT-based scores, VideoPhy benchmark and human evaluation. 
\end{abstract}
\vspace{-0.3cm}
\vspace{-1.2em}
\section{Introduction}
\label{sec:intro}
\vspace{-0.7em}

Recent advances in image-to-video (I2V) generation have achieved impressive fidelity and realism. The purely data-driven models~\citep{blattmann2023stable, openai2024sora, deepmind2025veo3, runway2024gen3alpha}, however, often lack fundamental understanding of real-world physics, yielding implausible dynamics and photonics, as noted by physics benchmarks~\citep{motamed2025generative, bansal2024videophy, meng2024towards}.
Several works aim to enhance the data-driven video models with physics awareness for general scenes from single images. Force prompting~\citep{gillman2025force} induces information of physical interactions into its model, while VLIPP~\citep{yang2025vlipp} employs vision-language models (VLM) to apply physics reasoning prior. However, these approaches fail in out-of-domain physics scenarios due to their strong reliance on data-driven models.

To obtain better generalization and accurate physics modeling, another line of works integrate explicit physics simulation into 3D Gaussian Splatting ~\citep{kerbl3Dgaussians, xie2024physgaussian} or learning differential simulators with video model priors \citep{zhang2024physdreamer, huang2025dreamphysics, liu2024physflow}. The physics-informed Gaussians are then rendered to generate physically plausible videos.
3D GS-based methods require multi-view inputs, which are difficult to obtain in practical settings.

Recently, PhysMotion~\citep{tan2024physmotion} and WonderPlay~\citep{li2025wonderplay} reconstruct 3D GS, apply physics simulation, and refine rendered outputs with video diffusion models.
PhysGen3D~\citep{chen2025physgen3d} extends
PhysGen~\citep{liu2024physgen} to 3D, modelling 3D physics from a single image via mesh reconstruction and Physically-Based Rendering (PBR).
PhysCtrl~\citep{physctrl2025} models physically simulated dynamics as 3D point trajectories using a diffusion model conditioned on material properties and applied forces. 
The aforementioned methods are limited to single objects since the adopted 3D reconstruction models are pre-trained on individual objects. The mesh representation in PhysGen3D struggles with modeling fabric and fluid dynamics where maintaining vertex connectivity is challenging. 
PerpetualWonder~\citep{zhan2026perpetualwonder} and PSIVG~\citep{foo2026physical}  reconstruct multi-object scenes and couple a physics simulator with a pretrained video diffusion model in a closed loop, achieving  physically realistic 4D representations.
However, PerpetualWonder includes multi-view video diffusion (Wan2.2) and per-step backward optimization of the visual-physical particles, that requires several hours of inference per scene, still sufferring from visual artifacts. PSIVG inherits the limited capability of its video model to adhere to the simulated dynamics.

In this work, we propose a novel pipeline, \textsc{3DPhysVideo}, for single image-to-3D dynamics video generation,
via two stages: single-view 3D reconstruction and simulation-guided video generation.
We propose \textit{Consistency-Guided Flow SDE}, which decomposes the predicted velocity of I2V flow model into denoising and consistency bias, enforces consistency to the conditional input, thereby repurposing the I2V model for both  stages. As a novel-view synthesizer, the I2V model generates a 360-degree orbit video with geometry guidance by unprojecting a single-view to the world coordinate using the point cloud reconstruction model VGGT~\citep{wang2025vggt}. 
Since the orbit video obtained by Consistency-Guided Flow SDE is world-consistent, we then obtain 3D geometry by simply unprojecting it. Note VGGT captures the distributions of generic scenes including multiple objects. 
Subsequently, using the I2V model as a simulation-guided video generator, we generate high-quality videos that follow 3D physics dynamics of various materials, including solids, fluids, and viscous substances, simulated via off-the-shelf physics solvers.
\textsc{3DPhysVideo} attains 3D geometry of generic scenes and multiple objects than single objects, offers high controllability on velocity, mass, and material properties, and generates photorealistic videos that obey simulated physics dynamics.
We validate our method through diverse experiments, both quantitatively and qualitatively via human evaluation, GPT-scores, and VideoPhy~\citep{bansal2024videophy}. We compare the proposed pipeline against state-of-the-art data-driven video generation models (e.g., Gen-3, Sora, VEO), and physics-aware approaches, including PhysGen3D, VLIPP, PhysCtrl, and PSIVG. We further assess our first and second stage isolated, i.e. single-view 3D reconstruction and simulation-guided video generation, in comparisons to relevant SOTA methods ~\citep{burgert2025gowiththeflow, xiao2025trajectory, bai2025recammaster, gu2025das} and ~\citep{ling2024motionclone, burgert2025gowiththeflow, li2025magicmotion, gu2025das}, respectively. Results show that our method significantly improves physical realism and semantic consistency thanks to Consistency-Guided Flow SDE, while maintaining competitive photorealism, particularly in challenging multi-object and fluid interaction scenarios. Our major contributions are: 
\begin{itemize} 
\item 
\vspace{-0.5em}
We propose \textsc{3DPhysVideo}, a novel training-free pipeline that combines pre-trained I2V model priors with physics simulation to generate 3D physically plausible videos from a single image. The I2V model serves as both a novel-view synthesizer and a simulation-guided video generation model.
\vspace{-0.2em}
\item   
We propose Consistency-Guided Flow SDE, a novel method that decomposes the predicted velocity of I2V flow model into denoising and consistency bias, yielding videos that are consistent with an input image and simulations, while maintaining  photorealism. 
\vspace{-0.5em}
\item 
Extensive experiments are conducted on more diverse scenes than prior explicit physical simulation-based methods~\citep{liu2024physgen, tan2024physmotion, chen2025physgen3d, physctrl2025, foo2026physical}. 
\end{itemize}

\vspace{-1.0em}
\section{Related Work}

\noindent \textbf{3D Physics Dynamics.}
3D Gaussian Splatting (3D GS) has been widely adopted as 3D scene representations. PhysGaussian~\citep{xie2024physgaussian} and subsequent works~\citep{huang2025dreamphysics, zhang2024physdreamer, cai2024gic, liu2024physflow, lin2025phys4dgen, lin2025omniphysgs, mittal2025uniphy} have assigned physics properties to Gaussian particles using Material Point Method (MPM)~\citep{stomakhin2013material, jiang2016materialpoint}. With recent advances in single-image 3D reconstruction, some methods directly predict 3D shapes~\citep{xu2024instantmesh} from a single view, while others adopt a two-step approach: synthesizing multi-view images~\citep{liu2023zero, shi2023mvdream} then reconstructing geometry~\citep{tang2024lgm, chen2025partgen}. 
Beyond object-centric scenarios, several studies~\citep{wang2025vggt, yu2025wonderworld, yu2024viewcrafter} extend to scene-level reconstruction. 
Building on single-view 3D reconstruction, existing works model 3D physical dynamics from a single image either by directly applying simulation at inference~\citep{chen2025physgen3d, tan2024physmotion, li2025wonderplay, zhan2026perpetualwonder, realwonder2026}  or leveraging simulated trajectories as supervision to learn physics-grounded motions as in PhysCtrl~\citep{physctrl2025}. Building on the inference-time simulation paradigm, 
PSIVG~\citep{foo2026physical} integrates Material Point Method (MPM) simulation directly into the video diffusion process, while PerpetualWonder ~\citep{zhan2026perpetualwonder} adopts a closed-loop framework alternating between physics simulation and video-based refinement.
These methods have shown the feasibility of generating physically realistic dynamic scenes from single-view inputs. See Sec.~\ref{sec:experiments} for comparisons. 

\noindent \textbf{Motion Conditioned Video Generation.}
A wide variety of methods have been explored for controllable video generation.  Optical flow has been used as conditioning signals enabling temporal coherence and global motion controls, but often lacks fine physical details~\citep{ni2023lfdm, liang2024movideo, chen2023mcdiff, burgert2025gowiththeflow}. Another line of work exploits trajectory and point-based control using point tracking models~\citep{SpatialTracker, xiao2025spatialtrackerv2}, which can flexibly handle both dense and sparse point conditions~\citep{geng2025motion, jeong2025track4gen, gu2025das}. Region and entity-based approaches ~\citep{wu2024draganything, li2025magicmotion, qiu2024freetraj} provide fine-grained controls via masks, bounding boxes, or landmarks, though they tend to overlook high-frequency motion details. Unified controllers ~\citep{zhang2025poe, wang2024motionctrl} aim to combine heterogeneous signals within a single framework. In parallel, training-free paradigms~\citep{jain2024peekaboo, ling2024motionclone} demonstrate the feasibility of repurposing pretrained video diffusion models without significant training overheads. 
3D-aware camera control has also gained attention, including~\citep{he2024cameractrl, yang2024direct, wu2025cat4d, xiao2025trajectory}, which improves multi-view consistency and enables explicit camera trajectory conditioning, while ReCamMaster~\citep{bai2025recammaster} extends this by re-rendering an input video along novel camera trajectories.
Beyond the trajectory or mask-based conditioning, the recent work on force prompting~\citep{gillman2025force} introduces physical forces as control signals for video generation, enabling both local point interactions and global effects. This demonstrates the generalisation of video models from limited synthetic training. 

\noindent \textbf{Flow-based Editing.}
Flow matching~\citep{lipman2022flow} has shown computational advantages over diffusion models
through straight-line ordinary differential equation (ODE) trajectories.
Recent works have explored various applications including text-driven image or video editing through inversion methods using score distillation~\citep{yang2024text}, rectified stochastic differential equation (SDE)~\citep{song2020score, rout2024semantic}, Taylor expansion solvers~\citep{wang2024taming}, and predictor-correction frameworks~\citep{jiao2025uniedit}. However, consistency to input images in editing via image-to-video flow matching remains under-explored.

\vspace{-1.0em}
\section{\textsc{3DPhysVideo}}
\label{sec:method}
\vspace{-0.7em}

Given a single RGB image $\mathbf{I} \in \mathbb{R}^{H \times W \times 3}$, our goal is to generate a physically plausible video $\mathbf{V}^{\text{sim}}$ through physics simulation. Our pipeline is designed to fully exploit the rich prior of a pretrained video generation model $G$~\citep{zhang2025framepack}. The first stage repurposes $G$  as a novel view synthesizer, producing an orbit video $\mathbf{V}^{\text{orb}}$ and corresponding 3D geometry from the input image $\mathbf{I}$ (Sec.~\ref{sec:3.1}). We then run physics simulation based on the 3D reconstruction, yielding simulated point trajectories $\mathcal{P} = \{\mathbf{P}_i\}_{i=1}^L$. In the second stage, we map these raw simulated trajectories $\mathcal{P}$ to a photorealistic and temporally coherent final video $\mathbf{V}^\text{sim}$  (Sec.~\ref{sec:3.2}). Both stages are made consistent with the conditional inputs by our novel Consistency-Guided Flow SDE, $\Phi_{\text{CF}}$, introduced in (Sec.~\ref{sec:4}).

\vspace{-0.7em}
\subsection{Stage 1: Single Image to 3D}  
\vspace{-0.3em}
\label{sec:3.1}
Obtaining 3D geometry from a single image is ill-posed due to invisible regions from the input view. 
Hence, it is critical to exploit strong priors of generative models. Unlike previous works~\citep{chen2025physgen3d, foo2026physical} that rely on object-level reconstruction~\citep{xu2024instantmesh} and thus discard inter-object information such as relative poses and positions, we instead leverage a generic video generation model G~\citep{zhang2025framepack} to reconstruct scenes including multi-objects. All masks needed in our pipeline are obtained using SAM2~\citep{ravi2024sam2}.
\vspace{-1.0em}
\paragraph{Point Cloud Unprojection.}
Given an image $\mathbf{I}$, we unproject a point cloud into the world coordinate using the point cloud reconstruction model VGGT~\citep{wang2025vggt}. 
Only foreground points of interest for physics modeling are retained.
\vspace{-1.0em}
\paragraph{Mesh Orbit Rendering.}
As shown in Fig.~\ref{fig:pipeline}, the unprojected point cloud provides correct geometry for visible regions from the input view. 
We then render the mesh by moving the camera along a 360-degree orbit trajectory. This produces a mesh orbit video $\{\mathbf{f}^{orb}_i\}_{i=1}^K$, which still has a significant number of missing pixels due to invisible regions and self-occlusions.
\vspace{-0.5em}
\paragraph{World-Consistent Orbit Video Generation.}
To obtain a complete world-consistent orbit video $\mathbf{V}^{\text{orb}}$ following the geometry guidance of {$\{\mathbf{f}^{orb}_i\}_{i=1}^K$, we repurpose the video model $G$ as a novel-view synthesizer. We achieve this using our Consistency-Guided Flow SDE, $\Phi_{\text{CF}}$ (Sec.~\ref{sec:4}) along with masking and inversion strategies~\citep{mokady2023null, wang2024taming} as
\begin{equation}
\mathbf{V}^\text{orb} = \Phi_{\text{CF}}(\{\mathbf{f}^{orb}_i\}_{i=1}^K,\mathbf{I},\mathbf{M}^\text{orb};G),
\label{eq:stage1}
\end{equation}
where $\mathbf{M}^{orb}$ is the video mask obtained by stacking the per-frame masks of $\{\mathbf{f}^\text{orb}_i\}_{i=1}^K$.
This mask helps preserve the geometry guidance regions, while enabling the use of video model $G$ that iteratively fills the empty regions to be semantically consistent with the input image $\mathbf{I}$.
\vspace{-0.5em}
\vspace{-0.5em}
\paragraph{3D Geometry Reconstruction.}
Since $\mathbf{V}^\text{orb}$ is world-consistent following the geometry guidance, we obtain the 3D geometry by simply unprojecting it back to the world coordinate using the same point cloud reconstruction model VGGT. By generating an orbit video that contains all foreground objects and then applying the scene-level reconstruction model~\citep{wang2025vggt}, our approach preserves inter-object relations—such as relative poses and positions. By contrast, the prior arts~\citep{chen2025physgen3d, foo2026physical} reconstruct each object independently and struggles to model multi-object interactions. 

\begin{figure*}[t]
    \centering
    \vspace{-0.5em}
\includegraphics[width=1.0\linewidth]{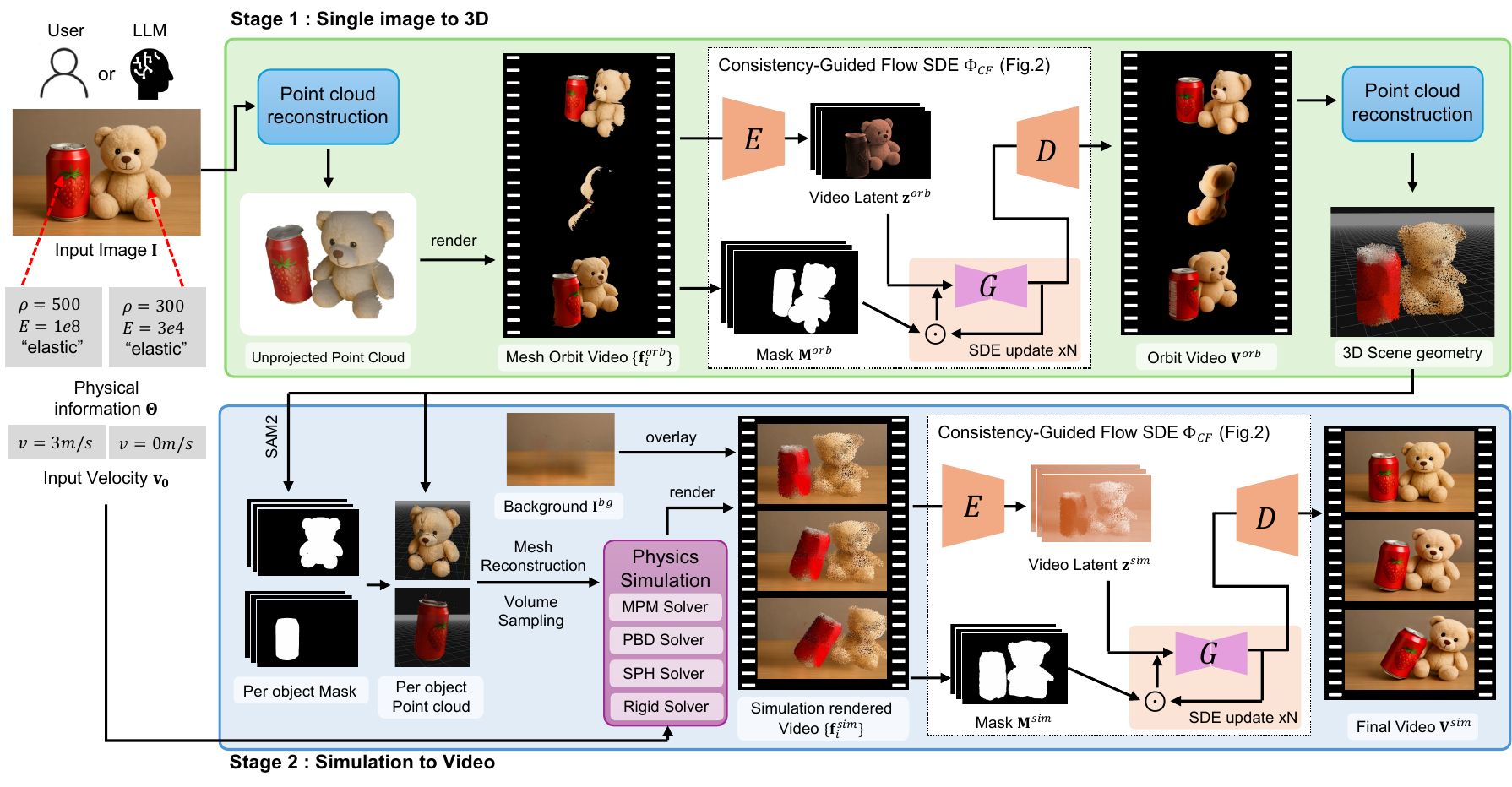}
 \caption{\textbf{Overall Pipeline.} Starting from a single image, \textsc{3DPhysVideo} reconstructs 3D geometry of multiple objects via 360-degree orbit video synthesis, then applies 3D point-based physics simulation to produce photorealistic videos from the simulation results.}
 \vspace{-1.5em}
    \label{fig:pipeline}
\end{figure*}

\vspace{-0.7em}
\subsection{Stage 2: Simulation to Video} 
\label{sec:3.2}
\vspace{-0.3em}

\paragraph{Physics Simulation \& Rendering.}
From the point cloud obtained in Sec.~\ref{sec:3.1}, we extract per-object point clouds for simulation using object segment masks. 
We convert it to a mesh and sample volumetric points to form simulation-ready point clouds $\{\mathbf{Q}^o_i\}$ with colors interpolated from the input point cloud. The ground plane $\pi$ is also obtained via RANSAC~\citep{fischler1981ransac}. Note the mesh reconstruction serves only for volumetric point sampling.


Following~\citep{chen2025physgen3d}, we obtain the initial velocity $\mathbf{v}_0$ and physical properties $\Theta$ through either automatic inference using GPT-5 or user specifications, including material parameters such as object elasticity, density, and surface friction coefficients required for realistic physics simulation.
With the obtained object-level point clouds $\{\mathbf{Q}^o_i\}$ and parameters $\pi$, $\mathbf{v}_0$ and $\Theta$, we run a physics simulator in Genesis~\citep{Genesis} to generate simulated output $\mathcal{P} = \mathtt{Sim}(\{\mathbf{Q}^o_i\}, \pi, \mathbf{v}_0, \Theta)$, where $\mathcal{P} = \{\mathbf{P}_i\}_{i=1}^L$ represents point trajectories that capture physically accurate deformations and motions.
We then render the raw simulated trajectories $\mathcal{P}$ using the point cloud renderer~\citep{pytorch3d_colored_points_tutorial}. These rendered frames are blended with the inpainted background image $\mathbf{I}^\text{bg}$, yielding the simulation-rendered video $\{\mathbf{f}^\text{sim}_i\}_{i=1}^L$.

\vspace{-0.5em}
\paragraph{Photorealistic Video Generation.}
The rendered video $\{\mathbf{f}^{sim}_i\}_{i=1}^L$ shows physically accurate deformations and motions. 
It however, lacks photorealism with unrealistic texture, missing shadows, and lighting effects. To address this, we repurpose the video model $G$ as a simulation-guided video generation model using $\Phi_{\text{CF}}$ with the masking and inversion strategies, similar to  Sec.~\ref{sec:3.1}, 
\begin{equation}
\mathbf{V}^{\text{sim}} = \Phi_{\text{CF}}(\{\mathbf{f}^{sim}_i\}_{i=1}^L,\mathbf{I},\mathbf{I}^\text{bg},\mathbf{M}^\text{sim};G),
\label{eq:stage2}
\end{equation}
where $\mathbf{M}^\text{sim}$ is the video mask obtained by concatenating the per-frame masks of $\{\mathbf{f}^\text{sim}_i\}_{i=1}^L$.
This mask with the background image enforces adherence to the simulated motion guidance and allows the video model $G$ to iteratively refine the appearance for consistency with the input image $\mathbf{I}$.
As a result, the final video $\mathbf{V}^{\text{sim}}$ is both physically faithful to the simulated dynamics and photorealistic. 

\vspace{-0.3em}
\subsection{Consistency-Guided Flow SDE $\Phi_{\text{CF}}$}
\label{sec:4}
\vspace{-0.3em}
A key contribution of our framework lies in leveraging the consistency-guided Flow SDE $\Phi_{\text{CF}}$ across two distinct stages. 
In Sec.~\ref{sec:3.1}, $\Phi_{\text{CF}}$ yields a world-consistent orbit video from an incomplete mesh orbit video.
In Sec.~\ref{sec:3.2}, $\Phi_{\text{CF}}$ translates a simulation-rendered video into a photorealistic video.
By reusing the same mechanism with different conditional guidances (geometry in the first stage and physics simulation in the second), our method unifies geometry reconstruction and physics-based animation under a single coherent framework. 
\vspace{-0.5em}
\paragraph{Initialization.}
\label{sec:4.1}
Consistency-guided Flow SDE is applied in the 
intermediate latent space at a diffusion step $t=\tau$ using the video model $G$.  
Initializing $\mathbf{z}_\tau$ in the manner tailored to a target task, our SDE enables general-purpose 
inductive bias optimization across a wide range of tasks.
For the world-consistent orbit video generation and simulation-to-video 
generation in Secs.~\ref{sec:3.1} and~\ref{sec:3.2},  
we leverage the masking and inversion to obtain $\mathbf{z}_\tau$ from 
the encoded latent $\mathbf{z}$ of the inputs of $\Phi_\text{CF}$, 
$\{\mathbf{f}^\text{orb}_i\}_{i=1}^K$ or 
$\{\mathbf{f}^\text{sim}_i\}_{i=1}^L$, in Eqs.~\ref{eq:stage1} 
and~\ref{eq:stage2} (see Sec.~\ref{app:algorithm} of the 
supplementary for details). 
For the other tasks in Sec.~\ref{app:consistency-guided-flow-sde-application} of the 
supplementary, such as coarse motion-conditioned video 
generation (e.g., bounding boxes, cut-and-drag) and non-visual inductive bias-guided optimization (e.g., text-alignment), 
a simple forward process 
$\mathbf{z}_\tau = (1-\tau) \cdot \mathbf{z} + \tau \cdot \boldsymbol{\epsilon}$ with $\boldsymbol{\epsilon} \sim \mathcal{N}(0, \mathbf{I})$ 
suffices. 
By replacing the random Gaussian noise from which a standard I2V model starts with $\mathbf{z}_\tau$ obtained from the input video, we translate the I2V model $G$ to video-to-video (V2V) tasks in a training-free manner.

\begin{figure*}[t]
  \centering
  \includegraphics[width=1.0\linewidth]{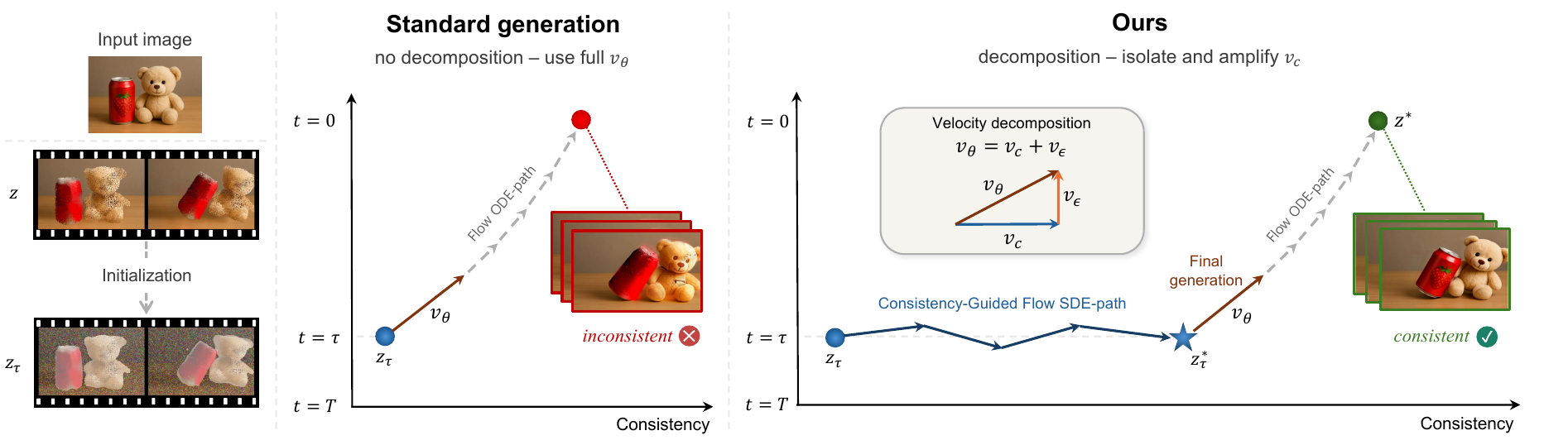}
  \vspace{-1.0em}
  \caption{\textbf{Consistency-Guided Flow SDE $\Phi_{\text{CF}}$.}
  Given an input video latent $\mathbf{z}$, we initialize a latent $\mathbf{z}_\tau$ at a diffusion step $t=\tau$. Since the standard generation process (left) follows the full velocity $v_\theta$, which includes the consistency bias $v_c$ and the denoising bias $v_\epsilon$, along the flow ODE-path, the brief exposure to $v_c$ leaves the result not aligned with the input image in texture, semantics, or other attributes. In contrast, our $\Phi_{\text{CF}}$ (right) decomposes $v_\theta = v_c + v_\epsilon$ and isolates $v_c$ to iteratively amplify the consistency bias at $t=\tau$ before final generation, yielding output $\mathbf{z}^{*}$ aligned with the input image.}
   
  \vspace{-1.5em}
  \label{fig:cf-sde}                    
\end{figure*}
\vspace{-0.5em}
\paragraph{Consistency-Guided Flow Stochastic Differential Equation (SDE).}
\label{sec:4.2}
The velocity prediction  $v_\theta$ of the I2V flow model $G$~\citep{lipman2022flow, zhang2025framepack} 
inevitably embodies both consistency bias (aligning the generated video with the input image) and denoising bias (moving toward the data manifold). 
Hence, a straightforward approach to generate a video from $\mathbf{z}_\tau$ is to follow the standard generation process guided by $v_\theta$. In practice, however, the resulting video appears inconsistent with the
input image (see Figs.~\ref{fig:cf-sde} and~\ref{fig:ablation}).
To achieve consistency-optimal video generation, we decompose $v_\theta$ into $v_c$, which captures the consistency bias and $v_\epsilon$, which captures the denoising bias:
\begin{equation}
v_\theta(\mathbf{z}_{t},\mathbf{z}_{I},t) = v_{c}(\mathbf{z}_{t},\mathbf{z}_{I},t) + v_{\epsilon}(\mathbf{z}_{t},\mathbf{z}_{I},t),
\label{eq:decomposed}
\end{equation}
where $\mathbf{z}_{I}$ is the latent of the input image $\mathbf{I}$. Our goal is to leverage $v_{c}$ at the intermediate diffusion step $t=\tau$ to optimize the video latent $\mathbf{z}_\tau$ into a consistency-optimal video latent $\mathbf{z}_{\tau}^*$ in Fig.~\ref{fig:cf-sde}.
By leveraging $v_{\epsilon}$ to maintain the original distribution $q = \mathcal{N}((1-\tau)\boldsymbol{\mu}, \tau^2 \mathbf{I})$ at $\tau$, we ensure that the enforced consistency does not lean toward either high-frequency details ($t\approx0$) or low-frequency components ($t\approx T$).
The objective is to find the target distribution $p^*$ that enables sampling $\mathbf{z}_{\tau}^*$ which maximizes $C(\cdot,\mathbf{z}_I)$ (the consistency with $\mathbf{z}_I$) while minimizing KL divergence~\citep{kullback1951information} with $q$, balanced by a regularization parameter $\beta$ as
\vspace{-0.8em}
\begin{equation}
p^* = \arg\max_{p} \; \mathbb{E}_{\mathbf{z}_\tau \sim p} \left[{C}(\mathbf{z}_{\tau},\mathbf{z}_I)\right] - \frac{1}{\beta}\mathbb{D}_{\mathrm{KL}}(p \,\|\, q).
\label{eq:objective}
\end{equation}

\vspace{-0.5em}
To solve this optimization, the consistency bias $v_{c}$ approximates $\nabla C(\mathbf{z}_{\tau},\mathbf{z}_I)$ to maximize the first term, while the denoising bias  $v_{\epsilon}$ is utilized to approximate the score function $\nabla_{\mathbf{z}_\tau} \log q(\mathbf{z}_\tau)$ for the KL divergence term.
Using these approximated gradients, we formulate an overdamped Langevin stochastic differential equation (SDE)~\citep{risken1989fokker, song2020score} to sample from the optimal distribution $p^*$ and discretize it via Euler-Maruyama method~\citep{kloeden1977numerical, kloeden1992numerical}.
Through this process with $v_c = v_\theta - v_\epsilon$ from our decomposition, we prove that at the specific choice $\beta = \frac{1-\tau}{\tau}$, the $v_\epsilon$ term completely cancels out. This enables us to achieve our consistency optimization objective using only the known $v_\theta$:

\vspace{-1.8em}
\begin{equation}
\begin{split}
    \mathbf{z}_{\tau}^{(n+1)}
    &= \left(1-\frac{\gamma}{\tau}\right)\mathbf{z}_{\tau}^{(n)}
    + \frac{1-\tau}{\tau}\gamma\ v_{\theta}(\mathbf{z}_{\tau}^{(n)},\mathbf{z}_{I},\tau) + \sqrt{2\gamma}\,\boldsymbol{\epsilon}^{(n)},
    \quad
    \boldsymbol{\epsilon}^{(n)}\sim\mathcal{N}(0,\mathbf{I}),
\end{split}
\label{eq:cf_formula}
\end{equation}

\vspace{-0.8em}
where $\gamma$ denotes the step size of discretization. 
Through $N$ iterations of this  update rule on $\mathbf{z}_{\tau}$, we obtain $\mathbf{z}^*_{\tau}$ semantically and photorealistically consistent with $\mathbf{z}_I$. 
Finally, we obtain $\mathbf{V}^\text{orb}$ or $\mathbf{V}^\text{sim} = \textit{D}(\textit{F}(\mathbf{z}_{\tau}^*))$ through the final generation \textit{F} and the decoder  \textit{D}.
The detailed derivation and algorithm are provided in Secs.~\ref{app:derivation} and~\ref{app:algorithm} of the supplementary.
\vspace{0.5em}

\vspace{-1.1em}
\section{Experiments}
\label{sec:experiments}
\vspace{-0.5em}
\noindent\textbf{Implementation Details.} 
In the proposed pipeline, we use Framepack~\citep{zhang2025framepack}, an auto-regressive DiT-based I2V flow model, which delivers SOTA performance and is relatively computationally light. Note our method is agnostic to I2V models. 
For physics simulation, we adopt the Genesis~\citep{Genesis}, which unifies several different physics solvers, including Material Point Method. We use MPM on the following experiments and other solvers as shown in Fig.~\ref{fig:teaser} (also see Sec.~\ref{app:simulation} of the supplementary) and set $\tau$ at 20 out of 25 inference steps, and we use $N$ = 10 and $\gamma$ = 0.2 for SDE optimization. All experiments were conducted on a single NVIDIA RTX 3090 GPU. 
Analysis of the SDE hyperparameters and the pipeline’s inference time breakdown are provided in Secs.~\ref{sec:appendix-hyperparam} and~\ref{sec:appendix-inference} of the supplementary material.  

\noindent\textbf{Compared Methods.}
We evaluate the proposed full pipeline (Stage~1 + Stage~2) against several strong baselines.  
FramePack~\citep{zhang2025framepack} serves as a baseline to isolate the effect of our physics-guided framework.  
Sora~\citep{openai2024sora}, Gen-3 Alpha~\citep{runway2024gen3alpha}, and Veo-3~\citep{deepmind2025veo3} are  state-of-the-art commercial video generation models, representing the data-driven methods.
Among explicit simulator-based methods, we compare
against two representative works: PhysGen3D~\citep{chen2025physgen3d}
(object-centric reconstruction; alongside
PhysMotion~\citep{tan2024physmotion} and
WonderPlay~\citep{li2025wonderplay}) and
PSIVG~\citep{foo2026physical} (multi-object reconstruction; alongside
PerpetualWonder~\citep{zhan2026perpetualwonder}). 
PerpetualWonder is excluded, as it did not generalize reliably to our evaluation scenarios despite taking several hours of per-scene optimization on H100 GPU, precluding meaningful comparisons. 

We also compare with VLIPP~\citep{yang2025vlipp}, a representative of methods that adapt pretrained video generation models (e.g., Force Prompting~\citep{gillman2025force}) to physically plausible motions without explicit simulations. Finally, we compare PhysCtrl~\citep{physctrl2025}, a diffusion-based method for physics-grounded single-object motion.
This part of comparisons are to methods that are applicable for the setting of single-image input, which is the central focus of our work.  The approaches of physics-integrated Gaussian Splatting~\citep{xie2024physgaussian, cai2024gic, lin2025omniphysgs, mittal2025uniphy} are not included, as they require multi-view inputs.  

For Stage~1 (single image to 3D), we compare against state-of-the-art camera-controlled video generation methods: Go-with-the-Flow~\citep{burgert2025gowiththeflow}, Trajectory Attention~\citep{xiao2025trajectory}, ReCamMaster~\citep{bai2025recammaster}, and Diffusion as Shader~\citep{gu2025das}, which represent warped-noise, attention-based, video-conditioned, and 3D tracking-based camera control, respectively.

\begin{figure*}[t]
    \centering
    
    \includegraphics[width=1.0\linewidth]{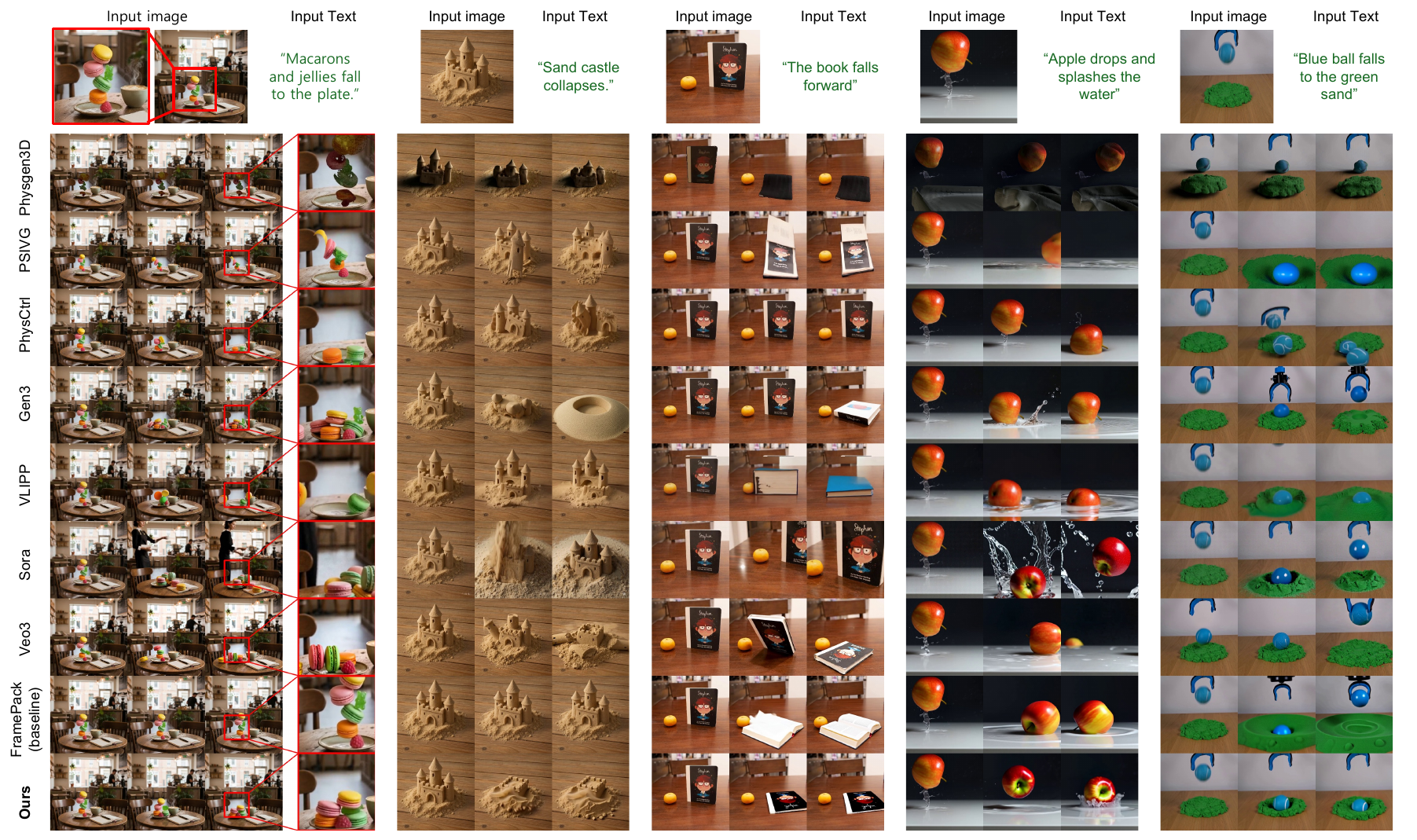}
    \vspace{-1.5em}
    \caption{Qualitative comparison across five representative scenarios. 
    From left to right: (i) several macarons and jellies falling onto a plate, 
    (ii) a sand castle collapsing, 
    (iii) a book falling forward,  
    (iv) an apple dropping into water, and 
    (v) a ball dropping to green sand. }
    \label{fig:qualitative_final2}
    \vspace{-1.3em}
\end{figure*}

\begin{figure}[!h]
    \centering
    \includegraphics[width=0.9\linewidth]
    {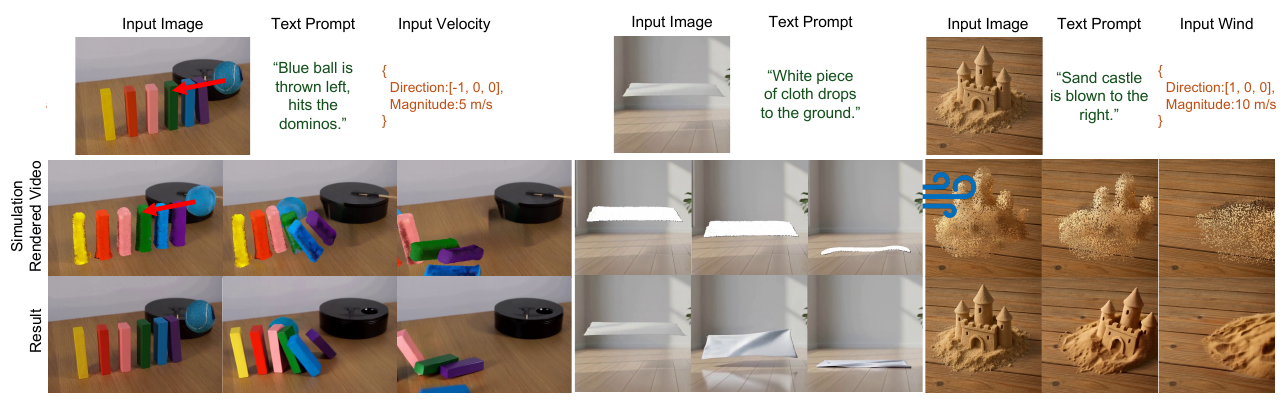}
    \vspace{-0.5em}
    \caption{
    Qualitative results on complex, non-local physical phenomena. Left: high-speed ball impact producing long-range domino cascades, made by composing Physics-IQ benchmark scenes. Middle: a piece of white cloth falling and deforming on the ground. Right: wind causing sand dispersion in the sand-castle scene.
    }
    \vspace{-2.0em}
    \label{fig:nonlocal}
\end{figure}

For Stage~2 (simulation to video), we compare against MotionClone~\citep{ling2024motionclone}, Go-with-the-Flow~\citep{burgert2025gowiththeflow}, MagicMotion~\citep{li2025magicmotion}, and Diffusion as Shader~\citep{gu2025das} which represent motion-prior, flow-based, mask-based conditioning, and 3D tracking-based, respectively. 

\noindent\textbf{Evaluation Metrics \& Datasets.}
Similar to~\citep{liu2024physgen, chen2025physgen3d, li2025wonderplay, physctrl2025, foo2026physical, zhan2026perpetualwonder}, we curate a dataset of ten diverse scenes. It spans a wider spectrum than the prior-arts, of materials such as rigid bodies, elastic objects, fluids, sand, and snow, paired with diverse physical phenomena, including global forces (e.g., wind), local forces (e.g., poking), and complex multi-object collisions. Note that no standard benchmark exists for video generation with explicit 3D physics simulation, unlike data-driven and LLM-guided methods, which can be evaluated on benchmarks such as Physics-IQ~\citep{motamed2025generative} (see Sec.~\ref{app:capability} of the supplementary). For assessing the physical realism, visual fidelity, and controllability of generated videos, we adopt GPT-5-based evaluation, scoring along three axes: Physical Realism (PhysR), Photorealism (PhotoR), and Semantic Consistency (Sem). We further conduct a user study with 20 participants on the same axes. For broader physical and visual assessment, we employ the VideoPhy benchmark~\citep{bansal2024videophy} to measure adherence to physical laws (denoted VPhy), and the VBench benchmark~\citep{huang2023vbench} for general video quality, reporting Aesthetic Quality (AQ), Background Consistency (BC), Imaging Quality (IQ), Motion Smoothness (MS), and Temporal Flickering (TF). Detailed evaluation protocols are provided in the supplementary material.

\newcolumntype{G}{>{\centering\arraybackslash}X}    
\newcolumntype{H}{>{\centering\arraybackslash\hsize=0.7\hsize}X}  
\newcolumntype{V}{>{\centering\arraybackslash\hsize=0.7\hsize}X}   

\begin{table*}[htbp]
\centering
\footnotesize
\renewcommand{\arraystretch}{0.95}
\vspace{-0.5em}
\caption{Quantitative comparisons by GPT-5 evaluation, VideoPhy Physical Commonsense Score (VPhy), and Human evaluation. GPT-5 scores are reported as mean$\pm$std.}
\vspace{-0.5em}
\label{tab:comprehensive_final}
\begin{tabularx}{\textwidth}{@{}l|*{3}{G}|V|*{3}{H}}
\toprule
\multirow{2}{*}{\textbf{Method}} & \multicolumn{3}{c|}{\textbf{GPT}} & \multirow{2}{*}{\textbf{VPhy~($\uparrow$)}} & \multicolumn{3}{c}{\textbf{Human}} \\
\cmidrule{2-4} \cmidrule{6-8}
 & \textbf{PhysR~($\uparrow$)} & \textbf{PhotoR~($\uparrow$)} & \textbf{Sem~($\uparrow$)} &  & \textbf{PhysR~($\uparrow$)} & \textbf{PhotoR~($\uparrow$)} & \textbf{Sem~($\uparrow$)} \\
\midrule
PhysGen3D & 0.179{\scriptsize$\pm$0.026} & 0.389{\scriptsize$\pm$0.027} & 0.241{\scriptsize$\pm$0.043} & 0.149 & 1.54 & 1.61 & 1.69 \\
PSIVG     & 0.371{\scriptsize$\pm$0.027} & 0.725{\scriptsize$\pm$0.020} & 0.428{\scriptsize$\pm$0.040} & 0.170 &   2.62  &  3.31   &  2.71   \\
PhysCtrl  & 0.334{\scriptsize$\pm$0.035} & 0.779{\scriptsize$\pm$0.022} & 0.276{\scriptsize$\pm$0.037} & 0.218 & 2.09 & 3.10 & 2.26 \\
VLIPP     & 0.574{\scriptsize$\pm$0.027} & 0.781{\scriptsize$\pm$0.019} & 0.579{\scriptsize$\pm$0.042} & \underline{0.244} & 1.95 & 2.67 & 2.52 \\
FramePack & 0.424{\scriptsize$\pm$0.038} & 0.805{\scriptsize$\pm$0.018} & 0.445{\scriptsize$\pm$0.050} & 0.221 & 2.17 & 3.19 & 2.39 \\
Gen3      & 0.557{\scriptsize$\pm$0.038} & 0.822{\scriptsize$\pm$0.015} & 0.528{\scriptsize$\pm$0.054} & 0.192 & 2.58 & 3.36 & 2.94 \\
Sora      & 0.629{\scriptsize$\pm$0.032} & 0.799{\scriptsize$\pm$0.018} & 0.489{\scriptsize$\pm$0.047} & 0.140 & 2.01 & 2.74 & 2.42 \\
Veo3       & \underline{0.725{\scriptsize$\pm$0.021}} & \textbf{0.882{\scriptsize$\pm$0.007}} & \underline{0.687{\scriptsize$\pm$0.039}} & 0.129 & \underline{3.16} & \underline{3.68} & \underline{3.47} \\
\textbf{\textsc{3DPhysVideo} (Ours)} & \textbf{0.727{\scriptsize$\pm$0.029}} & \underline{0.830{\scriptsize$\pm$0.011}} & \textbf{0.719{\scriptsize$\pm$0.040}} & \textbf{0.266} & \textbf{3.47} & \textbf{3.79} & \textbf{3.98} \\
\bottomrule
\end{tabularx}
\vspace{-2.0em}
\end{table*}

\vspace{-1.3em}
\subsection{Results}
\vspace{-0.7em}
\paragraph{Comparison with State-of-the-Art methods.}

We provide qualitative comparisons against baselines in Fig.~\ref{fig:qualitative_final2}. Given the same input image and text prompt, most of the baselines struggle to faithfully capture real-world physics.
Note, specifically, since PhysGen3D performs object-wise 3D reconstruction, it discards relative poses and spatial relations despite its additional optimization step.

As shown in the blue ball dropping scenario (the rightmost), PhysGen3D yields  inaccurate depth predictions placing the ball too far and low, leading to an unnatural simulation trajectory. Additionally, its mesh-based  rendering produces distorted textures and limited fluid dynamics, as observed in the apple scene (the fourth column). PhysCtrl, restricted to a single-object training setting, exhibits clear limitations in multi-object and liquid scenarios (see Sec.~\ref{app:more-results} in the supplementary for more results). PSIVG, despite explicitly integrating simulation into video diffusion, shows lower physical realism, as its reliance on a generated template video for 4D reconstruction lets the video model's prior override the simulator-derived trajectories. 

The data-driven models also exhibit physically implausible behaviors, such as objects suddenly disappearing and appearing, or moving unnaturally. For example, in the scenario of six macarons and jellies falling (the leftmost), Gen-3 and Sora hallucinate seven or nine, Veo3 removes one, and FramePack keeps them frozen in place. In contrast, our method consistently preserves the number of objects and appearance while generating physically plausible motions and realistic interactions.

\begin{wrapfigure}{r}{0.40\textwidth}
    \centering
    \vspace{1.5em}
    \includegraphics[width=\linewidth]{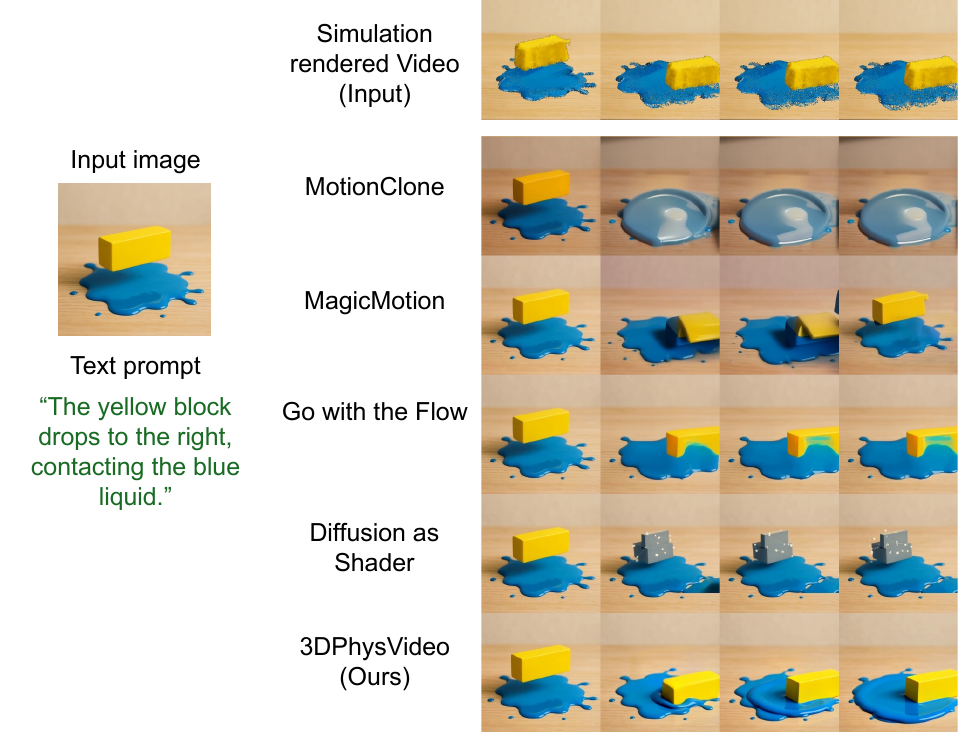}
    \caption{Qualitative comparison on Simulation to Video. }
    \vspace{-0.5em}
    \label{fig:step2_quali}
\end{wrapfigure}

Tab.~\ref{tab:comprehensive_final} summarizes the quantitative results. Our method achieves the highest overall scores in physical realism and semantic consistency. Meanwhile, our photorealism score remains competitive with state-of-the-art methods, demonstrating that improved physical plausibility does not compromise visual quality.

\vspace{-0.8em}

\paragraph{Complex and non-local physical phenomena.}
Beyond localized interactions, \textsc{3DPhysVideo} also handles large-scale, non-local physical scenarios (Fig.~\ref{fig:nonlocal}), such as multi-object collisions, cloth dynamics, and wind effects. In all cases, our method preserves the input appearance while generating coherent, globally coupled motion faithful to the simulated dynamics, demonstrating robustness in handling complex non-local physical effects. We further demonstrate additional applications of $\Phi_{\mathrm{CF}}$, including coarse motion-conditioned video generation and text alignment, in Sec.~\ref{app:consistency-guided-flow-sde-application} of the supplementary.

\vspace{-0.8em}

\subsection{Ablation Study}
\vspace{-0.3em}

\paragraph{Camera-controlled video generation.}

To demonstrate the 3D consistency of our generated orbit videos in Stage~1, we assess the camera trajectory accuracy of our method against state-of-the-art camera-controlled video generation methods in Tab.~\ref{table:stage1}. Following CameraCtrl~\citep{he2024cameractrl}, we measure RotErr and TransErr on both our curated 10 static scenes with orbit camera trajectories and 100 dynamic real-world video-trajectory pairs from ReCamMaster~\citep{bai2025recammaster} covering 10 distinct trajectory types (e.g., pan, tilt, zoom). While the baselines demand significant training cost, \textsc{3DPhysVideo} outperforms all baselines on static scenes and achieves competitive accuracy even on dynamic scenes. See qualitative results in Sec.~\ref{app:stage1_qualitative} of the supplementary.

\begin{table}[htbp]
\vspace{-1.3em}
\caption{Stage 1 comparison on camera-controlled video generation: training cost, RotErr, TransErr, and 109-frame inference time. ``A100-h'' and ``H800-h'' denote GPU-hours on the respective hardware.}
\label{table:stage1}
\centering
\fontsize{7.0}{8.3}\selectfont
\setlength{\tabcolsep}{7.8pt}
\renewcommand{\arraystretch}{1.0}
\begin{tabular}{@{}l|c|cc|cc|c}
\toprule
\multirow{2}{*}{\textbf{Method}} & \multirow{2}{*}{\textbf{Train Cost}} & \multicolumn{2}{c|}{\textbf{Static Scenes}} & \multicolumn{2}{c|}{\textbf{Dynamic Scenes}} & \multirow{2}{*}{\textbf{Inf. Time~($\downarrow$)}} \\
\cmidrule(lr){3-4} \cmidrule(lr){5-6}
 & & \textbf{RotErr~($\downarrow$)} & \textbf{TransErr~($\downarrow$)} & \textbf{RotErr~($\downarrow$)} & \textbf{TransErr~($\downarrow$)} & \\
\midrule
Go-with-the-Flow     & 7,680 A100-h & \underline{1.2774}          & \underline{0.293}          & 0.6722          & 0.0031          & 15.6 min          \\
Trajectory Attention & 24 A100-h    & 2.3033          & 0.398          & 0.6194          & 0.0171          & \textbf{8.4 min}  \\
ReCamMaster          & 136K videos  & 1.6826          & 0.350          & 0.6443          & 0.0044          & 16.1 min          \\
Diffusion as Shader  & 576 H800-h   & 1.7370          & 0.326          & \textbf{0.6100} & \textbf{0.0025} & 26.7 min          \\
\midrule
\textbf{\textsc{3DPhysVideo} (Ours)} & \textbf{Training-free} & \textbf{0.2598} & \textbf{0.116} & \underline{0.6119} & \underline{0.0029} & \underline{12.9 min} \\
\bottomrule
\end{tabular}
\vspace{-2.5em}
\end{table}
\begin{table}[htbp]
\vspace{0.3em}
\caption{Stage 2 comparison on motion-conditioned video generation: training cost, GPT-based and VBench metrics, and 109-frame inference time. ``A100-h'' and ``H800-h'' denote GPU-hours on the respective hardware.}
\label{table:stage2}
\centering
\setlength{\tabcolsep}{3.5pt}
\renewcommand{\arraystretch}{1.1}
\resizebox{\linewidth}{!}{
\begin{tabular}{@{}l|c|ccc|ccccc|c}
\toprule
\multirow{2}{*}{\textbf{Method}} & \multirow{2}{*}{\textbf{Train Cost}} & \multicolumn{3}{c|}{\textbf{GPT}} & \multicolumn{5}{c|}{\textbf{VBench}} & \multirow{2}{*}{\textbf{\shortstack{Inf.\\Time~($\downarrow$)}}} \\
\cmidrule(lr){3-5} \cmidrule(lr){6-10}
 & & \textbf{PhysR~($\uparrow$)} & \textbf{PhotoR~($\uparrow$)} & \textbf{Sem~($\uparrow$)} & \textbf{AQ~($\uparrow$)} & \textbf{BC~($\uparrow$)} & \textbf{IQ~($\uparrow$)} & \textbf{MS~($\uparrow$)} & \textbf{TF~($\uparrow$)} & \\
\midrule
MotionClone         & Training-free   & 0.379{\scriptsize$\pm$0.058} & 0.642{\scriptsize$\pm$0.069} & 0.422{\scriptsize$\pm$0.091} & \underline{0.551} & 0.865          & 0.612          & 0.980          & 0.972          & 20.4 min          \\
MagicMotion         & Fine-tuned         & 0.418{\scriptsize$\pm$0.078} & 0.585{\scriptsize$\pm$0.047} & 0.593{\scriptsize$\pm$0.095} & 0.531          & 0.921          & 0.608          & 0.990          & 0.986          & 33.4 min          \\
Go-with-the-Flow    & 7,680 A100-h    & 0.515{\scriptsize$\pm$0.076} & 0.755{\scriptsize$\pm$0.038} & 0.587{\scriptsize$\pm$0.088} & 0.540          & 0.940          & 0.622          & \underline{0.995} & \underline{0.992} & \underline{15.6 min} \\
Diffusion as Shader & 576 H800-h      & \underline{0.574{\scriptsize$\pm$0.054}} & \underline{0.830{\scriptsize$\pm$0.027}} & \underline{0.649{\scriptsize$\pm$0.053}} & 0.536 & \underline{0.941} & \underline{0.630} & \textbf{0.996} & \underline{0.992} & 26.7 min \\
\midrule
\textbf{\textsc{3DPhysVideo} (Ours)} & \textbf{Training-free} & \textbf{0.709{\scriptsize$\pm$0.071}} & \textbf{0.883{\scriptsize$\pm$0.024}} & \textbf{0.784{\scriptsize$\pm$0.117}} & \textbf{0.580} & \textbf{0.942} & \textbf{0.635} & \textbf{0.996} & \textbf{0.993} & \textbf{12.9 min} \\
\bottomrule
\end{tabular}
}
\vspace{-2.0em}
\end{table}

\vspace{-0.3em}

\paragraph{Motion-conditioned video generation.}
The purpose of using a physics simulator is to produce physically realistic videos. This, however, is not achieved when the video models fail to faithfully follow the simulated motions.
In Fig.~\ref{fig:step2_quali} and Tab.~\ref{table:stage2}, we report the comparison results of our Stage 2 against state-of-the-art motion-conditioned video generation methods to evaluate how accurately each model follows the simulated motion. For a fair comparison, the same simulated motion, input image, and text prompt are provided to all methods. Our method precisely follows the simulated motion and achieves the best performance, demonstrating not only superior physical plausibility but also consistently strong visual fidelity.



\begin{wrapfigure}{r}{0.45\textwidth}
    \centering
    \vspace{-1.2em}
    \includegraphics[width=\linewidth]{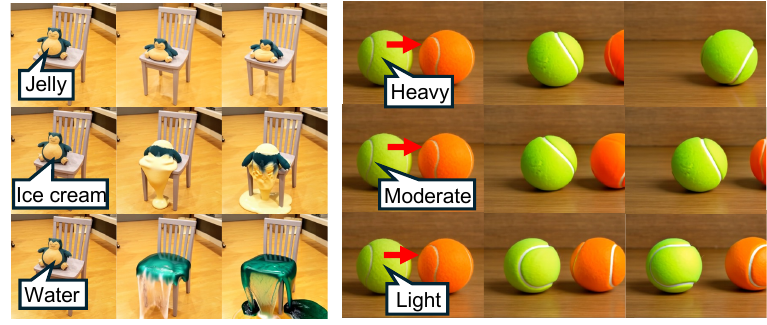}
    \caption{Qualitative results under different physical properties.}
    \label{fig:dynamic}
    \vspace{-1.0em}
\end{wrapfigure}

\noindent\textbf{Physical dynamics.}
Fig.~\ref{fig:dynamic} shows that our framework allows flexible control over different input physical parameters, generating results that faithfully reflect the specified properties. We vary the material (jelly, ice cream, water) and the mass under the same velocity. Since we use video priors to synthesize faithful videos from simulation results, this enables handling out-of-domain physics scenarios (e.g., Snorlax with ice cream or water-like properties) where data-driven generation methods fail.


    

\begin{wrapfigure}{r}{0.45\textwidth}
    \centering
    \includegraphics[width=\linewidth]{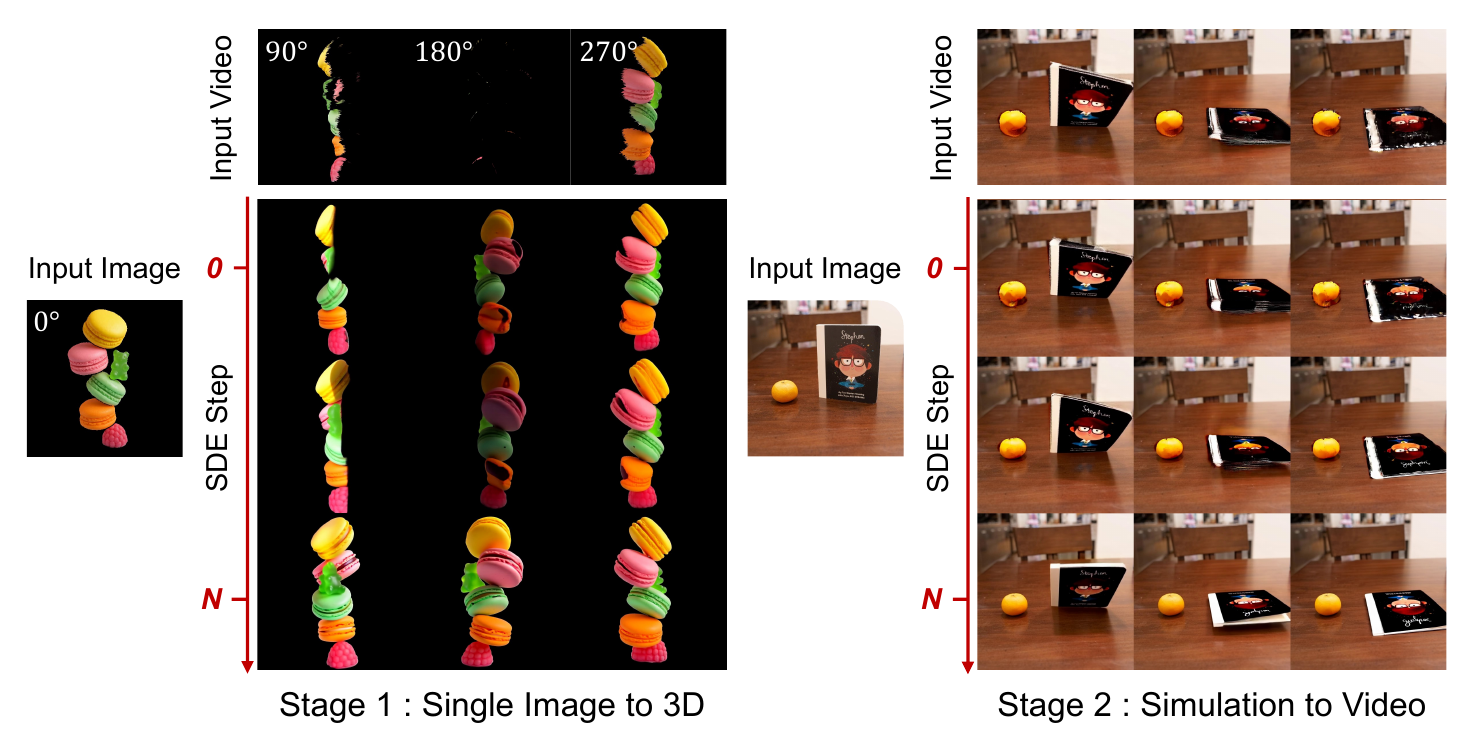}
    \caption{Ablation on the number of Consistency-Guided Flow SDE steps.}
    \label{fig:ablation}
\end{wrapfigure}

\vspace{-0.8em}
\paragraph{Effects of SDE optimization.} Fig.~\ref{fig:ablation} shows the effectiveness of our Consistency-Guided Flow SDE by varying the number of SDE optimization steps in Stage~1 (left) and Stage 2 (right). With zero SDE step (i.e., standard generation), unseen regions in the orbit views remain incomplete due to missing pixels in the mesh orbit rendering. As the number of SDE steps increases, the orbit views progressively fill in these missing areas, producing a geometrically consistent video that faithfully follows the geometry guidance while remaining semantically aligned with the input image. For Stage 2, 
the input simulation-rendered video of the falling book scene is gradually refined with SDE step $= 0$ to $N$.

\vspace{-1.0em}
\section{Conclusions}
\vspace{-0.5em}
We introduced a novel training-free pipeline for generating a physically plausible video from a single input image. 
This work represents an important step forward by demonstrating how off-the-shelf video generation models can be repurposed, without a significant computational overhead of training, as a novel-view synthesizer and simulator-guided renderer, for single-view 3D reconstruction and photorealistic video generation.
The core of our framework, the proposed Consistency-Guided Flow SDE, leverages the model's inherent consistency bias and can also be applied as a general-purpose bias enforcer for applications that require different inductive biases, such as text alignment.
Future work includes exploring additional applications and extension to more complex physics dynamics for real-world modeling.


\bibliographystyle{plainnat}   
\bibliography{example_paper}

\clearpage
\appendix

\section{Teaser Figure Details}
\label{app:teaser_details}

The teaser figure (Fig.~\ref{fig:teaser}) shows four physical phenomena---a hanging robe, a duck floating on water, a Franka Panda striking foam, and a rising steam plume---each generated by our full pipeline: 3D scene reconstruction (Sec.~\ref{sec:3.1}), physics simulation Genesis with parameters following Sec.~\ref{app:simulation}, and simulation-to-video refinement via Consistency-Guided Flow SDE $\Phi_{\text{CF}}$ (Secs.~\ref{sec:3.2} and~\ref{sec:4}). The four resulting videos are then combined into the shared bathroom scene for visualization.

\vspace{-0.3em}
\paragraph{Simulation setup.}
We drive four phenomena using Genesis~\citep{Genesis}: a duck floating on water, a hanging robe, a Franka Panda striking foam, and a rising steam plume. Common parameters are listed in Tab.~\ref{tab:sim-params}.
\vspace{-0.1em}
\paragraph{Duck on water.}
Water is simulated as SPH liquid (\texttt{SPH.Liquid}; kinematic viscosity $\mu{=}5{\times}10^{-3}$, particle size $1.3{\times}10^{-2}$, regular-lattice sampling) inside a rigid bath container, coupled to the duck through Genesis' SPH--rigid coupling such that particles are displaced as the body moves through them. The duck itself is driven kinematically along a hand-crafted 3D trajectory: at each step, its pose is written via \texttt{set\_qpos} and DoF velocities are zeroed.
\vspace{-0.1em}
\paragraph{Hanging robe.}
A PBD cloth on a 2-manifold silhouette mesh, with the back edge pinned via \texttt{fix\_particles} to an invisible hanger rod. After a $1$\,s warm-up drape, a small sinusoidal wind impulse applied every $8$ frames maintains visible sway against PBD damping.
\vspace{-0.1em}
\paragraph{Bubble.}
Approximately $20$k MPM Snow particles ($E{=}8{\times}10^{4}$, $\nu{=}0.32$, $\rho{=}40$) on a $128^3$ grid with $\Delta t{=}2{\times}10^{-3}$\,s and $25$ substeps. An inverted Franka Panda with stiffened PD gains ($\mathbf{k}_p, \mathbf{k}_v$ in Tab.~\ref{tab:sim-params}) follows an analytic strike trajectory $z(t) = z_0 + h - (h+d)\,s$, where $s = \tfrac{1}{2}\left(1 - \cos 2\pi n t\right)$, with inverse kinematics re-solved at every substep so that the controller never operates on a stale setpoint.
\vspace{-0.1em}
\paragraph{Steam.}
MPM Snow particles ($E{=}10^{2}$, $\nu{=}0.10$, $\rho{=}15$) with reversed gravity. At each step we apply height-growing lateral diffusion, a tangential vortex curl $c\,e^{-r/0.20}\sin(\omega t)\,(-\sin\theta,\cos\theta)$, an optional horizontal wind, and damping of $\dot{z}$ above $z{=}0.7$\,m. A \emph{recycle} variant teleports particles past $z{=}0.85$ back to the source location, enabling continuous emission.
\vspace{-0.1em}
\paragraph{Refinement back to our pipeline.}
After running these per-phenomenon simulations, the resulting point trajectories are rendered as in Sec.~\ref{sec:3.2} and passed through our Consistency-Guided Flow SDE $\Phi_{\text{CF}}$ (Sec.~\ref{sec:4}) to produce the photorealistic frames shown in Fig.~\ref{fig:teaser}. 

\section{Applications of Consistency-Guided Flow SDE}
\label{app:consistency-guided-flow-sde-application}

\paragraph{General Motion-Controlled Video Generation.}

With a simple forward process as initialization, our SDE extends to general motion-conditioned video generation in a training-free manner.
As shown in Fig.~\ref{fig:rebuttal_gwtf}, $\Phi_{\text{CF}}$ generates realistic dynamic motion following the coarse motion prior of the driving video while naturally adapting corrupted backgrounds to be consistent with the input image.

\begin{figure}[!h]
    \centering
    \includegraphics[width=1.0\linewidth]{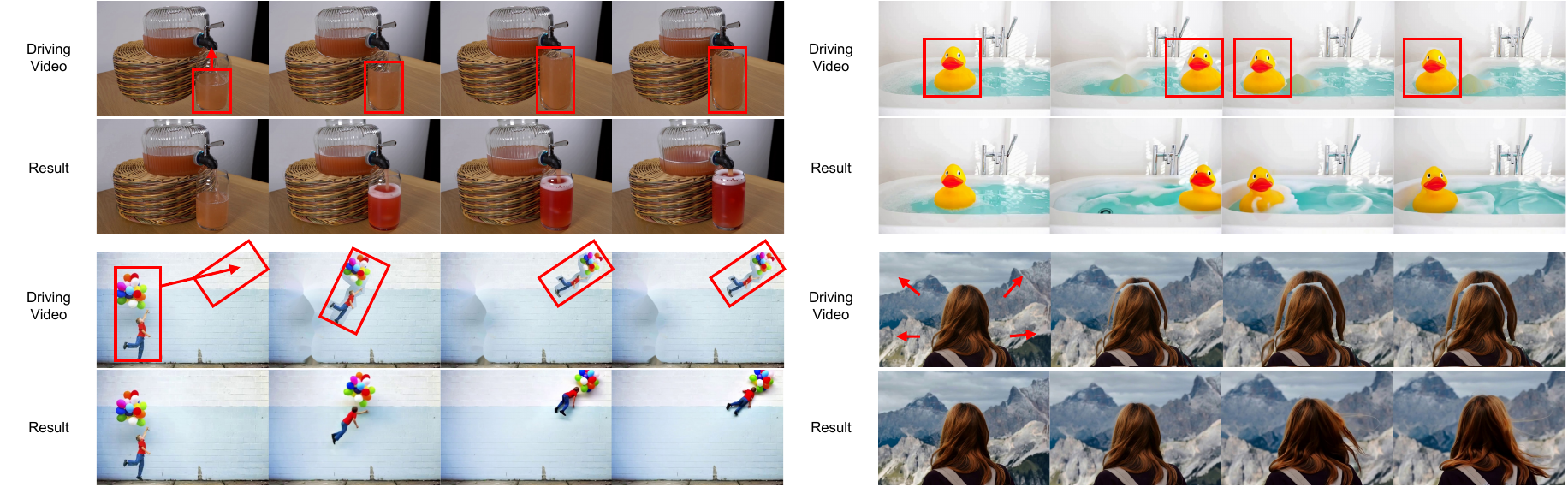}
    \caption{Qualitative results of Consistency-Guided Flow SDE given coarse motion priors. The juice scene (top left) uses a VLIPP-style driving video~\citep{yang2025vlipp} from the Physics-IQ benchmark; the remaining three use Cut-and-Drag driving videos from Go-with-the-Flow~\citep{burgert2025gowiththeflow}.}

    \label{fig:rebuttal_gwtf}
\end{figure}

\paragraph{General-Purpose Inductive Bias Optimization.}


Our SDE optimization generalizes beyond image consistency to other inductive biases acquired during training, such as text alignment. By providing an initial video misaligned with a target text prompt, we obtain a text-alignment bias $v_{\text{text}} = v_{\theta} - v_{\epsilon}$ that approximates $\nabla_{\mathbf{z}_\tau} \text{TextAlignment}(\mathbf{z}_{\tau}, \mathbf{z}_{\text{text}})$, analogously to the consistency bias derivation in Sec.~\ref{app:derivation}, and use it as the score term in the SDE---without any explicit text-alignment metric. As shown in Fig.~\ref{fig:non-visual}, this enables both progressive refinement of detailed text prompts that naive inference fails to fully reflect (left) and style-shifting text-guided video editing (right).

\begin{figure}[!h]
    \centering
    \includegraphics[width=1.0\linewidth]{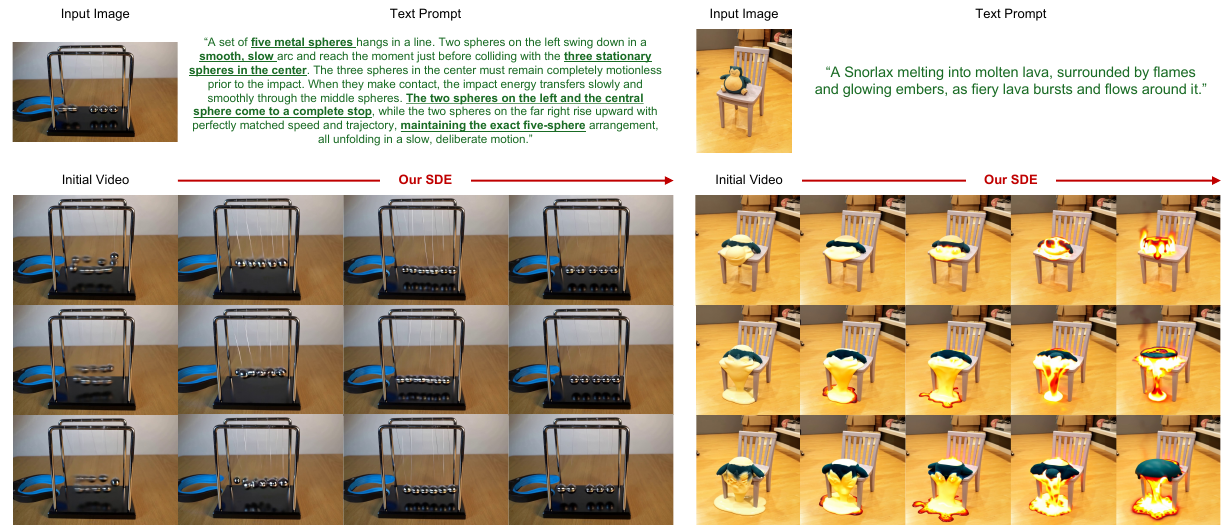}

    \caption{Qualitative results of our SDE with text alignment bias. Left: For a detailed text prompt, the initial video generated by FramePack~\citep{zhang2025framepack} from the prompt is progressively refined through our SDE to increasingly align with the detailed text prompt (note the \textbf{\underline{bold}} text). Right: An initial video from \textsc{3DPhysVideo} showing a snorlax melting like ice cream is edited to melt into burning lava according to the text prompt.}

    \label{fig:non-visual}
\end{figure}

\section{Scope Comparison with Explicit Simulation-Based Methods}
\label{app:capability}
To clarify the scope of physical phenomena and reconstruction settings supported by simulator-based video generation methods, we provide a comparison in Tab.~\ref{tab:capability}. As shown in the ``Curated'' column, all simulator-based approaches---including ours---necessarily evaluate on curated scene sets rather than fixed benchmarks such as Physics-IQ~\citep{motamed2025generative}. This is because (i) rendering simulation outputs requires reconstructed 3D geometry that is not available in pre-recorded benchmark videos, and (ii) different methods adopt different solvers that target different materials and scenes. Within this common setting, \textsc{3DPhysVideo} jointly reconstructs multiple objects in a single pass while supporting a wider range of physical dynamics than the relevant prior arts.

\begin{table}[h]
    \centering
    \footnotesize
    \setlength{\tabcolsep}{4pt}
    \renewcommand{\arraystretch}{0.95}
    \caption{Scope comparison of explicit simulator-based video generation methods. ``Curated'' indicates evaluation on a self-curated scene set (a common practice for simulation-based methods); ``Multi-obj.'' indicates the method handles scenes with multiple objects; ``Joint'' indicates joint reconstruction of multiple objects in a single pass, preserving inter-object spatial relationships; remaining columns indicate supported physical phenomena.}
    \label{tab:capability}
    \begin{tabular}{l|c|c|cc|cccc}
    \toprule
    \textbf{Method} & \textbf{Venue} & \textbf{Curated} & \textbf{Multi-obj.} & \textbf{Joint} & \textbf{Fluid} & \textbf{Granular} & \textbf{Wind} & \textbf{Cloth} \\
    \midrule
    PAC-NeRF         & ICLR'23    & \cmark &        &        & \cmark & \cmark &        &        \\
    PhysGaussian     & CVPR'24    & \cmark & \cmark &        & \cmark & \cmark & \cmark &        \\
    PhysDreamer      & ECCV'24    & \cmark &        &        &        &        & \cmark & \cmark \\
    PhysGen          & ECCV'24    & \cmark & \cmark &        &        &        &        &        \\
    OmniPhysGS       & ICLR'25    & \cmark & \cmark &        & \cmark & \cmark & \cmark & \cmark \\
    PhysGen3D        & CVPR'25    & \cmark & \cmark &        &        &        &        &        \\
    PhysCtrl         & NeurIPS'25 & \cmark &        &        &        & \cmark & \cmark &        \\
    WonderPlay       & ICCV'25    & \cmark & \cmark &        & \cmark & \cmark & \cmark & \cmark \\
    PSIVG            & CVPR'26    & \cmark & \cmark & \cmark &        &        &        &        \\
    \midrule
    \textbf{\textsc{3DPhysVideo} (Ours)} & -- & \cmark & \cmark & \cmark & \cmark & \cmark & \cmark & \cmark \\
    \bottomrule
    \end{tabular}
\end{table}
\section{Qualitative Results of Camera-Controlled Video Generation}
\label{app:stage1_qualitative}

Although our method primarily targets 3D reconstruction via static-scene orbit videos in Stage~1, the proposed Consistency-Guided Flow SDE $\Phi_{\text{CF}}$ itself generalizes beyond this setting. Given a source video and a camera trajectory as conditional input, it enables general camera-controlled video generation for both dynamic and static scenes. We provide qualitative results on dynamic scenes with diverse camera trajectories (Figs.~\ref{fig:stage1_qual_2}--\ref{fig:stage1_qual_5}), as well as on static scenes under full 360° orbital trajectories that match our Stage~1 reconstruction setting (Figs.~\ref{fig:stage1_qual_orbit_2}--\ref{fig:stage1_qual_orbit_1}).

\begin{figure}[htbp]
    \centering
    \includegraphics[width=\linewidth]{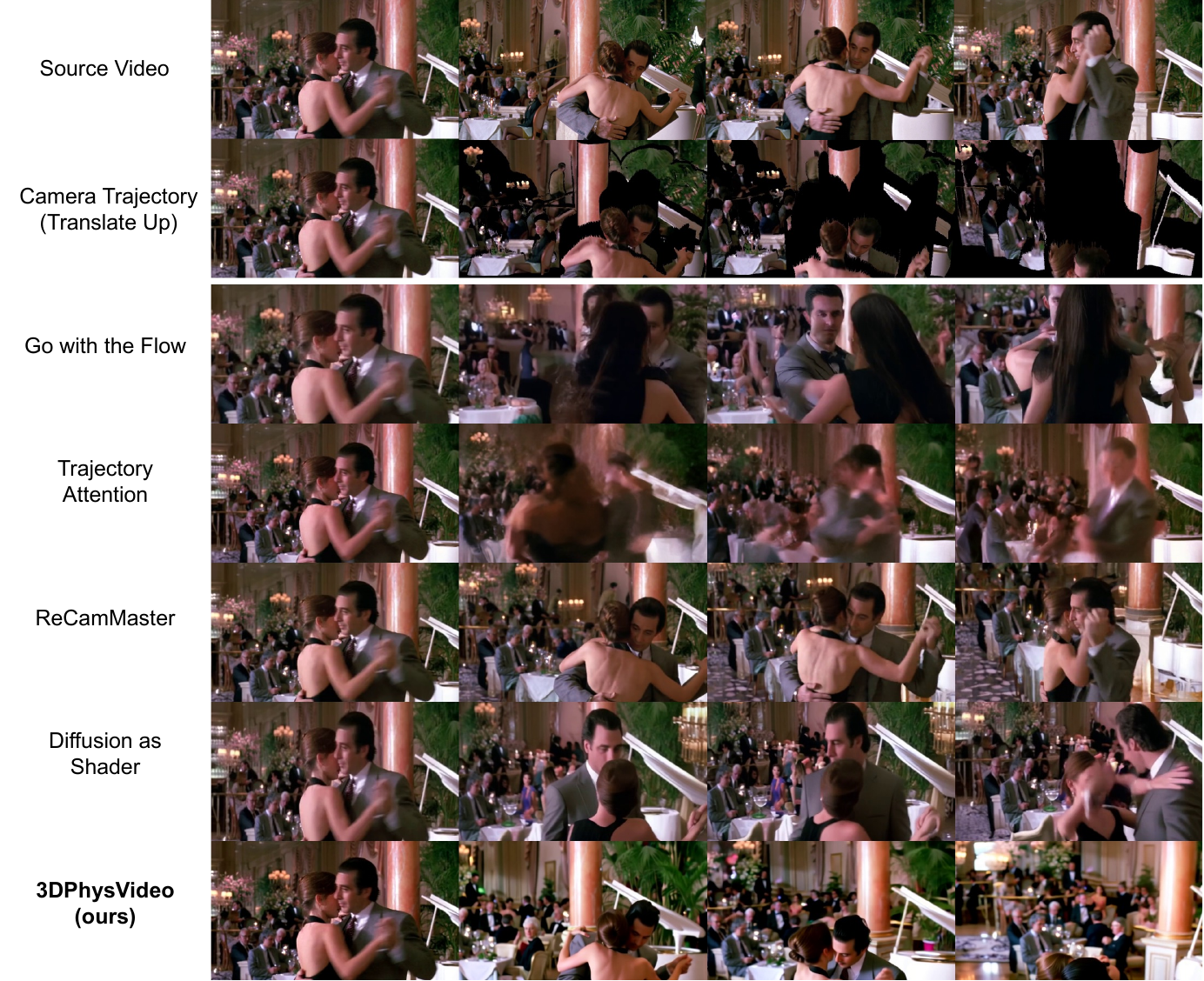}
    \caption{Qualitative result on the \emph{Translate Up} trajectory.}
    \label{fig:stage1_qual_2}
\end{figure}

\begin{figure}[htbp]
    \centering
    \includegraphics[width=\linewidth]{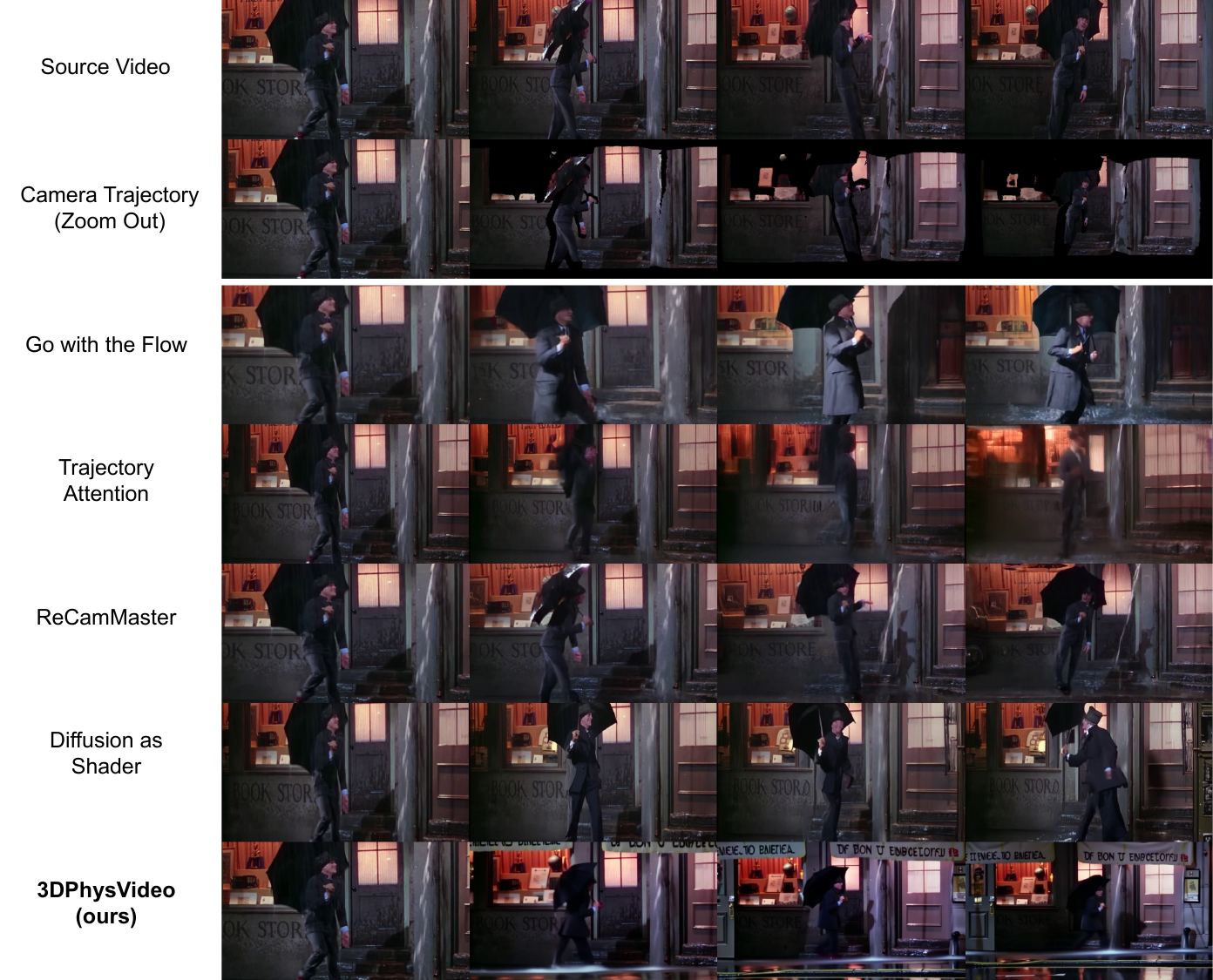}
    \caption{Qualitative result on the \emph{Zoom Out} trajectory.}
    \label{fig:stage1_qual_4}
\end{figure}

\begin{figure}[htbp]
    \centering
    \includegraphics[width=\linewidth]{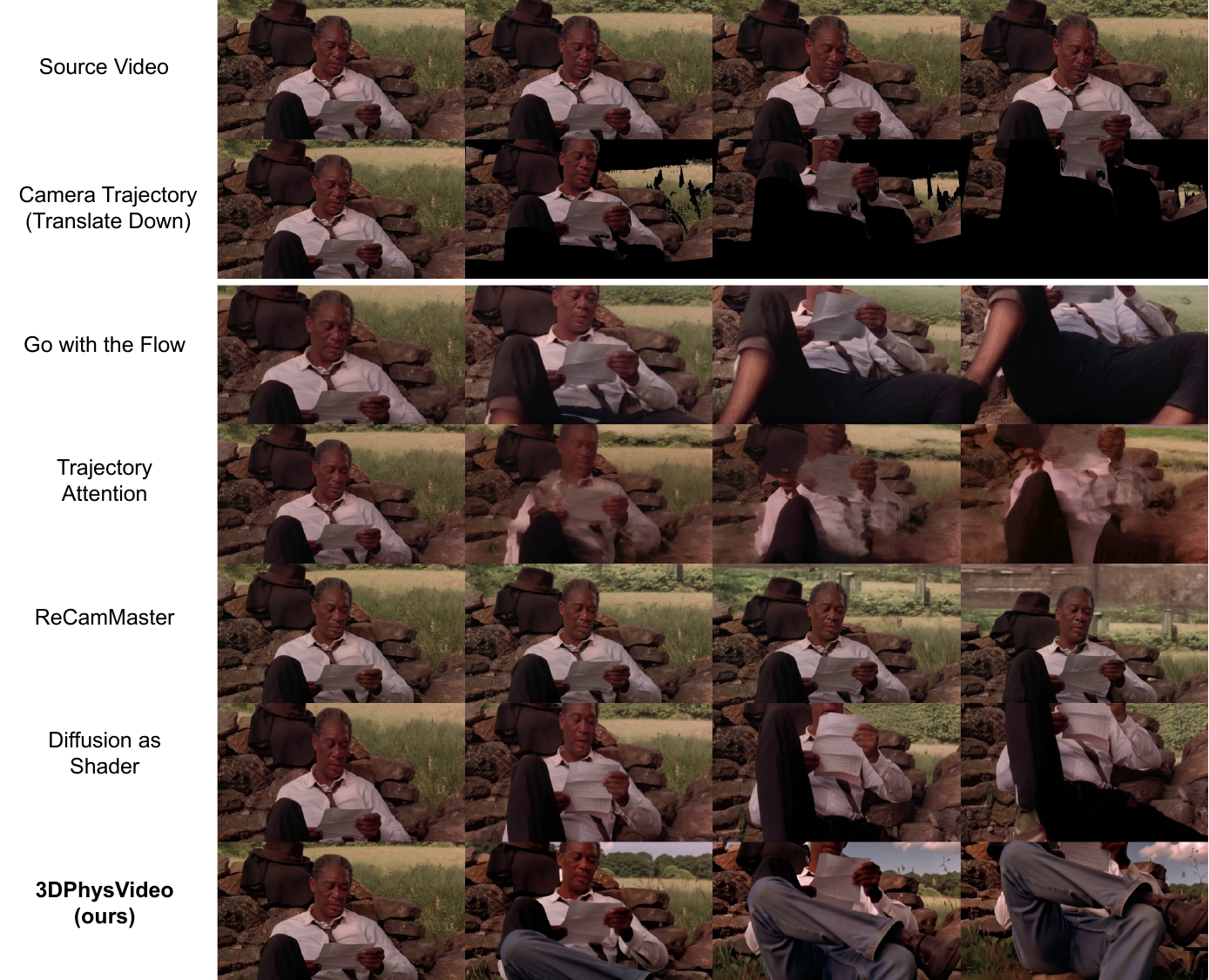}
    \caption{Qualitative result on the \emph{Translate Down} trajectory.}
    \label{fig:stage1_qual_5}
\end{figure}

\begin{figure}[htbp]
    \centering
    \includegraphics[width=\linewidth]{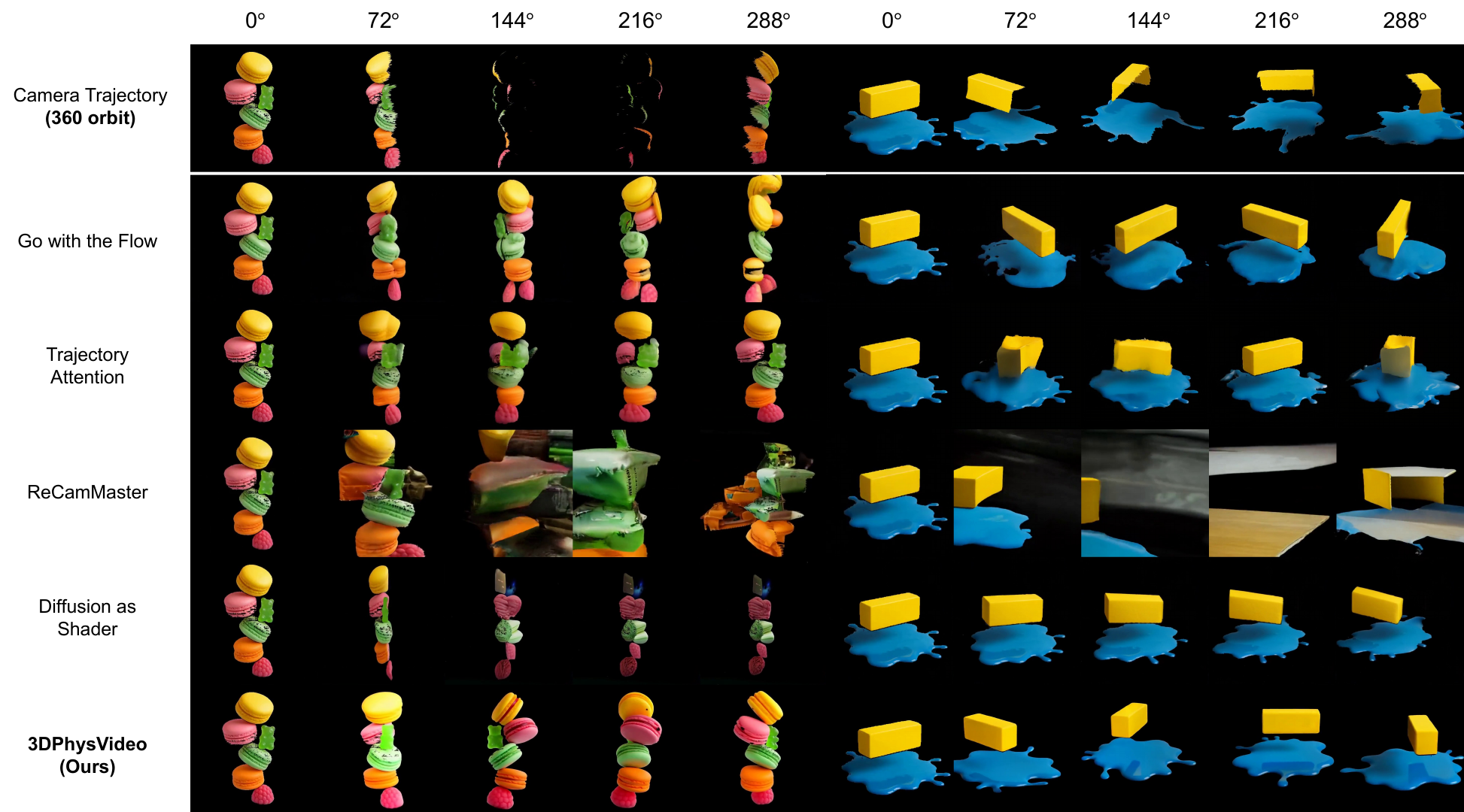}
    \caption{Qualitative results on a static scene under a 360° orbital camera trajectory.}
    \label{fig:stage1_qual_orbit_2}
\end{figure}

\begin{figure}[htbp]
    \centering
    \includegraphics[width=\linewidth]{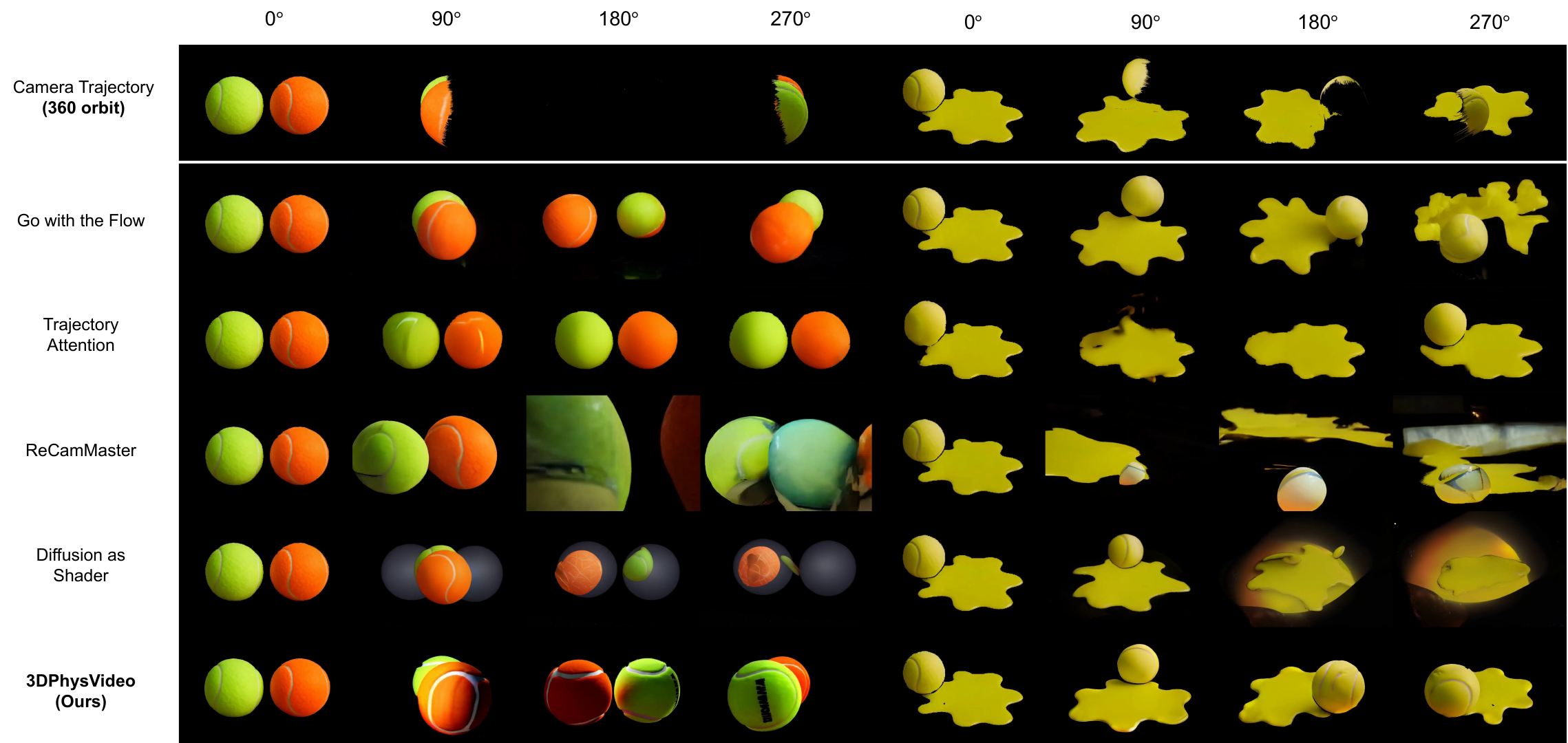}
    \caption{Qualitative results on a static scene under a 360° orbital camera trajectory.}
    \label{fig:stage1_qual_orbit_1}
\end{figure}

\FloatBarrier

\section{Detailed derivations for Consistency-Guided Flow SDE}
\label{app:derivation}
\vspace{1.0em}
To derive Eq.~\ref{eq:cf_formula}, we start from our objective in Eq.~\ref{eq:objective}:
\begin{equation}
p^* = \arg\max_{p} \; \mathbb{E}_{\mathbf{z}_\tau \sim p} \left[{C}(\mathbf{z}_{\tau},\mathbf{z}_I)\right] - \frac{1}{\beta}\mathbb{D}_{\mathrm{KL}}(p \,\|\, q).
\end{equation}
By expanding the KL divergence term and rewriting the objective in its minimizing form, we can rewrite it as:
\vspace{0.3em}
\begin{equation}
\arg\min_p \mathbb{E}_{\mathbf{z}_\tau \sim p}\left[\frac{1}{\beta} \log \frac{p(\mathbf{z}_\tau)}{q(\mathbf{z}_\tau)} - C(\mathbf{z}_\tau, \mathbf{z}_I)\right] = \arg\min_p \mathbb{E}_{\mathbf{z}_\tau \sim p}\left[\log \frac{p(\mathbf{z}_\tau)}{q(\mathbf{z}_\tau)} - \beta C(\mathbf{z}_\tau, \mathbf{z}_I)\right].
\end{equation}
\vspace{0.3em}
Following the exponential tilting result for KL-constrained objectives in \citep{rafailov2023direct}), the optimal distribution is:
\vspace{0.3em}
\begin{equation}
p^*(\mathbf{z}_\tau) = \frac{q(\mathbf{z}_\tau)\,\exp\bigl(\beta C(\mathbf{z}_\tau, \mathbf{z}_I)\bigr)}{\displaystyle\int q(\mathbf{z}_\tau')\,\exp\bigl(\beta C(\mathbf{z}_\tau', \mathbf{z}_I)\bigr)\,d\mathbf{z}_\tau'}.
\end{equation}
\vspace{0.3em}
Taking the score of this optimal distribution:
\vspace{0.3em}
\begin{equation}
\nabla_{\mathbf{z}_\tau} \log p^*(\mathbf{z}_\tau) = \beta\,\nabla_{\mathbf{z}_\tau} C(\mathbf{z}_\tau, \mathbf{z}_I) + \nabla_{\mathbf{z}_\tau} \log q(\mathbf{z}_\tau), 
\end{equation}
\vspace{0.3em}
this score function directly gives us the drift term for the overdamped Langevin SDE~\citep{risken1989fokker, song2020score}:
\begin{equation}
d\mathbf{z}_k = \bigl[\beta\,\nabla_{\mathbf{z}_\tau} C(\mathbf{z}_\tau, \mathbf{z}_I) + \nabla_{\mathbf{z}_\tau} \log q(\mathbf{z}_\tau)\bigr]\,dk + \sqrt{2}\,d\mathbf{W}_k,
\label{eq:app_sde}
\end{equation}
\vspace{0.3em}
where \(\mathbf{W}_k\) is a standard Wiener process. 
\vspace{0.3em}
To implement this, we apply our decomposed velocity models in Eq.~\ref{eq:decomposed}. We approximate the $
\nabla_{\mathbf{z}_\tau}C(\mathbf{z}_\tau, \mathbf{z}_I)$ using our consistency bias model $v_c = v_\theta - v_\epsilon$ and approximate the score function using the denoising bias model $v_\epsilon$ based on the flow ODE structure:
\vspace{0.3em}
\begin{equation}
\nabla_{\mathbf{z}_\tau} C(\mathbf{z}_\tau, \mathbf{z}_I) \approx v_c = v_\theta - v_\epsilon, 
\quad \nabla_{\mathbf{z}_\tau} \log q(\mathbf{z}_\tau) \approx -\frac{\mathbf{z}_\tau - (1-\tau)(\mathbf{z}_\tau+\tau v_\epsilon)}{\tau^2}.
\end{equation}
\vspace{0.3em}
By substituting these approximations into Eq.~\ref{eq:app_sde}:
\vspace{0.3em}
\begin{equation}
d\mathbf{z}_k = \bigl[\beta\,(v_\theta - v_\epsilon) - \frac{\mathbf{z}_\tau - (1-\tau)(\mathbf{z}_\tau+\tau v_\epsilon)}{\tau^2}\bigr]\,dk + \sqrt{2}\,d\mathbf{W}_k.
\end{equation}
\vspace{0.3em}
Applying the Euler-Maruyama discretization~\citep{kloeden1977numerical, kloeden1992numerical}. with step size $\gamma$ gives us the update rule:
\vspace{0.3em}
\begin{equation}
\begin{aligned}
\mathbf{z}_{\tau}^{(n+1)} 
&= \mathbf{z}_{\tau}^{(n)} + \bigl[\beta\,(v_\theta - v_\epsilon) - \frac{\mathbf{z}_{\tau}^{(n)} - (1-\tau)(\mathbf{z}_{\tau}^{(n)}+\tau v_\epsilon)}{\tau^2}\bigr]\,\gamma + \sqrt{2\,\gamma}\;\boldsymbol{\epsilon}^{(n)} \\[0.3em]
&= (1-\frac{\gamma}{\tau})\mathbf{z}_{\tau}^{(n)}
+ \beta\gamma\ v_{\theta}
- (\beta-\frac{1-\tau}{\tau})\gamma v_\epsilon
+ \sqrt{2\gamma}\,\boldsymbol{\epsilon}^{(n)},
\quad
\boldsymbol{\epsilon}^{(n)}\sim\mathcal{N}(0,\mathbf{I}).
\end{aligned}
\label{eq:general_sde}
\end{equation}
\vspace{0.5em}
This discretized iterative formula ensures convergence to the optimal distribution $p^*(\mathbf{z}_\tau)$ that satisfies our objective in Eq.~\ref{eq:objective}, as established by the exponential tilting result and the convergence properties of Euler-Maruyama method. 
Notably, when we choose $\beta = \frac{1-\tau}{\tau}$, the coefficient of $v_\epsilon$ vanishes: $(\beta - \frac{1-\tau}{\tau})\gamma = 0$
, causing the $v_\epsilon$ terms to completely cancel out. This yields a simplified final update rule in Eq.~\ref{eq:cf_formula} that depends only on the known velocity model $v_\theta$, eliminating the need for explicit knowledge of the denoising bias model $v_\epsilon$ while maintaining convergence to the desired optimal distribution.
\vspace{1.0em}
\section{Algorithm and Analysis}
\label{app:algorithm}
A standard I2V model generates a video from a random Gaussian noise sample (with the same dimensionality as the target video latent) conditioned on an input image. In the proposed Consistency-Guided Flow SDE $\Phi_{\text{CF}}$ replaces this random Gaussian noise with $\mathbf{z}_\tau$, obtained by initializing from the input video, and uses it as the starting latent (Alg.~\ref{alg:algorithm1}, lines 1--8). In a training-free manner, this enables the I2V model to be applied to the video-to-video (V2V) tasks i.e. taking an inconsistent input video conditioned on an input image and yielding a consistent output video.

We provide the implementation of our $\Phi_{\text{CF}}$.
The Alg.~\ref{alg:algorithm1} takes $\{\mathbf{f}^\text{orb}_i\}_{i=1}^K$ or $\{\mathbf{f}^\text{sim}_i\}_{i=1}^L$ as the input rendered video $\{\mathbf{f}_i\}$ and $\mathbf{M}^\text{orb}$ or $\mathbf{M}^\text{sim}$ as the video mask $\mathbf{M}$, and outputs $\mathbf{V}^\text{orb}$ or $\mathbf{V}^\text{sim}$ as the final video $\mathbf{V}$.
Note that only the two lines (7, 12) differ between the two stages.

\begin{algorithm}[ht]
\caption{Consistency-Guided Flow SDE $\Phi_{\text{CF}}$}
\label{alg:main}
\textbf{Require:} An input image $\mathbf{I}$, input rendered video $\{\mathbf{f}_i\}$, video mask $\mathbf{M}$, background image $\mathbf{I}^\text{bg}$, SDE target timestep $\tau$, SDE optimization iterations $N$, one-step flow model generation process $F(\cdot, v_\theta, t)$, one-step flow model inversion process $F^{-1}(\cdot, v_\theta, t)$, encoder $E(\cdot)$, decoder $D(\cdot)$
\vspace{0.5ex}
\begin{spacing}{1.2} 
\footnotesize
\begin{algorithmic}[1]    
  \State $(\mathbf{z}, \mathbf{z}_I, \mathbf{z}^\text{bg}) \gets E(\{\mathbf{f}_i\},\mathbf{I},\mathbf{I}^\text{bg})$ \Comment{encodes each input to its corresponding latent}
  \vspace{0.5ex}
  \State $\mathbf{z}_{0}^{\text{inv}} \gets \mathbf{z}$
  \For{$t=0\ldots \tau-1$}
    \State $\mathbf{z}_{t+1}^\text{inv} \gets F^{-1}(\mathbf{z}_{t}^\text{inv},\,v_{\theta}(\mathbf{z}_{t}^\text{inv},\mathbf{z}_{I},t), t)$ 
  \EndFor
  \State $\boldsymbol{\epsilon}\sim\mathcal{N}(0,\mathbf{I})$
  \State $\mathbf{z}_{\tau}^{\text{noisy}} \gets \begin{cases} 
  (1-\tau)\cdot \mathbf{z}+\tau \cdot \boldsymbol{\epsilon} & \text{(stage 1)} \\
  (1-\tau)\cdot \mathbf{z}^{\text{bg}}+\tau \cdot \boldsymbol{\epsilon} & \text{(stage 2)}
  \end{cases}$
  \State $\mathbf{z}_\tau\gets \mathbf{M}\cdot \mathbf{z}_{\tau}^{\text{inv}}+(1-\mathbf{M})\cdot \mathbf{z}_{\tau}^{\text{noisy}}$
  \For{$n=0\ldots N-1$}
    \State $\boldsymbol{\epsilon}^{(n)}\sim\mathcal{N}(0,\mathbf{I})$
    \State $\mathbf{\hat z}_\tau^{(n+1)}\gets(1-\frac{\gamma}{\tau})\mathbf{z}_\tau^{(n)}
      +\frac{1-\tau}{\tau}\gamma v_{\theta}(\mathbf{z}_\tau^{(n)},\mathbf{z}_{I},\tau)
      +\sqrt{2\gamma}\,\boldsymbol{\epsilon}^{(n)}$ \Comment{Eq. \ref{eq:cf_formula}}
    \State $\mathbf{z}_\tau^{(n+1)} \gets \begin{cases} 
    \mathbf{M}\cdot \mathbf{z}_\tau^{(n)}+(1-\mathbf{M})\cdot \mathbf{\hat z}_\tau^{(n+1)} & \text{(stage 1)} \\
    \mathbf{M}\cdot \mathbf{\hat z}_\tau^{(n+1)}+(1-\mathbf{M})\cdot \mathbf{z}_\tau^{(n)} & \text{(stage 2)}
    \end{cases}$
  \EndFor\hspace{0.2em} $\mathbf{z}_{\tau}^{*}\gets \mathbf{z}_\tau^{(N)}$
  \For{$t=\tau\ldots1$}
    \State $\mathbf{z}_{t-1}^*\gets F(\mathbf{z}_{t}^*,v_{\theta}(\mathbf{z}_{t}^*,\mathbf{z}_{I},t), t)$
  \EndFor\hspace{0.2em} $\mathbf{z}^*\gets \mathbf{z}_{0}^*$
  \State $\mathbf{V} \gets D(\mathbf{z}^*)$ \Comment{decodes the latent to video}
  \State \textbf{Output:} Final video $\mathbf{V}$
\end{algorithmic}
\end{spacing}
\label{alg:algorithm1}
\end{algorithm}

To clarify the algorithm's unified approach, we provide detailed explanations for both use cases, highlighting the minimal but crucial differences between the two stages:

\noindent\textbf{For Stage 1.} The mesh orbit video $\{\mathbf{f}^\text{orb}_i\}_{i=1}^K$ provides accurate geometry for regions visible in the input image, and our goal is to fill the remaining empty regions while preserving this geometric guidance. To this end, line 12 keeps the inverted latent $\mathbf{z}_\tau^\text{inv}$ in the masked region $\mathbf{M}$ and updates only the unmasked regions, ensuring the SDE optimization fills empty regions while strictly following the geometry guidance.

\noindent\textbf{For Stage 2.} The simulation-rendered video $\{\mathbf{f}^\text{sim}_i\}_{i=1}^L$ provides accurate motion but unrealistic appearance, so the goal is to refine appearance while preserving motion. Two simple modifications suffice: (i) line 7 uses $\mathbf{z}^\text{bg}$ instead of $\mathbf{z}$, preventing the simulated motion from spreading into static regions; (ii) line 12 updates only the masked region $\mathbf{M}$ (opposite to Stage 1), so motion regions are refined into photorealistic appearance while unmasked static regions retain natural backgrounds---shadows, lighting, and ambient motion---from the initial noise $\boldsymbol{\epsilon}$.

\vspace{3.0em}

\section{Physical Simulation Parameters}
\label{app:simulation}
\vspace{1.0em}
The forward physics pass employs various solvers (rigid body, MPM, SPH, PBD), and each solver requires specific physical parameter settings for reasonable simulation. We provide a comprehensive list of these parameters, along with their default values, in Tab.~\ref{tab:sim-params}. In practice, following the common practice of prior physics-aware video generation methods~\citep{li2025wonderplay, zhan2026perpetualwonder}, these parameters are initially estimated using a Vision-Language Model and are subject to optional manual fine-tuning to ensure physically plausible simulation results.

\begin{table}[t]
\centering
\small
\setlength{\tabcolsep}{6pt}
\renewcommand{\arraystretch}{1.05}
\caption{Genesis simulation parameters used in our experiments. Unlisted parameters use the Genesis default values.}
\label{tab:sim-params}
\begin{tabular}{lc}
\toprule
\textbf{Parameter} & \textbf{Value} \\
\midrule
\multicolumn{2}{l}{\textit{General simulation}} \\
\quad Step time $\Delta t$
  & $4\times10^{-3}$ \\
\quad Sub-steps number
  & $10$ \\
\quad Sampled particle size
  & $1.3\times10^{-2}$ \\
\quad Gravity
  & $(0,\,0,\,-9.81)$ \\
\midrule
\multicolumn{2}{l}{\textit{Rigid body solver}} \\
\quad Friction coefficient
  & $0.2$ \\
\quad Coupling friction
  & $0.3$ \\
\quad Object density $\rho$
  & $180\ \mathrm{kg/m^3}$ \\
\quad Joint stiffness $k_p$
  & $[9000,9000,7000,7000,4000,4000,4000,200,200]$ \\
\quad Joint damping $k_v$
  & $[700,700,600,600,350,350,350,15,15]$ \\
\midrule
\multicolumn{2}{l}{\textit{SPH solver}} \\
\quad Liquid kinematic viscosity $\mu$
  & $5\times10^{-3}$ \\
\quad Particle size
  & $1.3\times10^{-2}$ \\
\quad Sampling
  & regular lattice \\
\midrule
\multicolumn{2}{l}{\textit{MPM solver}} \\
\quad Grid density
  & $128$ \\
\quad Young's modulus $E$
  & $8\times10^{4}$ \\
\quad Poisson's ratio $\nu$
  & $0.32$ \\
\quad Material density $\rho$
  & $40\ \mathrm{kg/m^3}$ \\
\midrule
\multicolumn{2}{l}{\textit{PBD solver}} \\
\quad Cloth stretch compliance
  & $1\times10^{-7}$ \\
\quad Cloth bending compliance
  & $1\times10^{-5}$ \\
\bottomrule
\end{tabular}
\vspace{0.3em}
\end{table}

\vspace{0.6em}
\section{Evaluation Protocol}
\subsection{Human Evaluation Protocol}
\label{appendix:human_eval}
\vspace{0.5em}
To complement the automatic metrics, we conducted a human evaluation following the protocol introduced in PhysGen3D. A total of 20 participants were recruited to assess the quality of generated videos. The evaluation was performed on 10 representative scenes, each rendered by 8 different methods, resulting in 70 videos in total. For each video, participants were asked to answer three questions corresponding to three quality dimensions: physical realism, photorealism, and semantic consistency.
\vspace{0.3em}
At the beginning of the study, participants were provided with the following instruction:
\vspace{0.6em}
\begin{quote}
\emph{We want to evaluate the quality of the generated video. You will be asked to assess it from the three perspectives: physical realism, photorealism, and semantic consistency.}
\end{quote}
\vspace{1.6em}
The three evaluation criteria were described in detail as follows:
\vspace{1.0em}
\begin{itemize}
    \item \textbf{Physical Realism} measures how realistically the video follows physical rules. Participants were asked to consider whether the video represents physical properties such as elasticity and friction, and whether the movements and interactions of objects behave plausibly and consistently with real-world expectations.
    \vspace{0.3em}
    \item \textbf{Photorealism} evaluates the visual fidelity of the video, including whether there are visual artifacts or discontinuities, and whether lighting, shadow, texture, and material details closely resemble real-world appearances.
    \vspace{0.3em}
    \item \textbf{Semantic Consistency} examines how well the generated video aligns with the provided input text and reference image.
\end{itemize}

\vspace{1.0em}
Each video was rated by answering the following three questions:  
1) \emph{The video is physically realistic.}  
2) \emph{The video is photorealistic.}  
3) \emph{The video is consistent with the input image and input text ``\{text prompt\}''.}  
\vspace{0.3em}
Responses were collected on a 5-point Likert scale (1 = strongly disagree, 5 = strongly agree). The order of videos was randomized across participants to mitigate ordering bias.
\vspace{0.3em}

\vspace{1.0em}
\subsection{GPT-based Evaluation Protocol}
\label{app:gpt-eval-protocol}
For completeness, we include the details of the GPT-based evaluation protocol, which we directly adopt from PhysGen3D~\citep{chen2025physgen3d}.  
In this protocol, GPT is prompted to assess generated videos along three axes: \emph{Physical Realism}, \emph{Photorealism}, and \emph{Semantic Consistency}.  
The evaluation is performed on evenly sampled frames from each video, together with the original input image and the motion instructions.  

\vspace{1.0em}
Specifically, GPT is instructed with the following template:  

\vspace{1.0em}
\begin{quote}
I would like you to evaluate the quality of \{num\_videos\} generated videos based on the following criteria: physical realism, photorealism, and semantic consistency.  

The evaluation will be based on \{num\_frames\} evenly sampled frames from each video.  
Given the original image and the following instructions: '\{instructions\}', please evaluate the quality of each video on the three criteria mentioned above.  

Note that:  

Physical Realism measures how realistically the video follows the physical rules and whether the video represents real physical properties like elasticity and friction. To discourage completely stable video generation, we instruct respondents to penalize such cases.  

Photorealism assesses the overall visual quality of the video, including the presence of visual artifacts, discontinuities, and how accurately the video replicates details of light, shadow, texture, and materials.  

Semantic Consistency evaluates how well the content of the generated video aligns with the intended meaning of the text prompt.  

Please provide the following details for each video, scores should be ranging from 0--1, with 1 to be the best: \{score\_lines\}  

Note that your output should strictly follow the above format, with a `;` after each score.  
Do not give any other explanations.
\end{quote}

\vspace{1.0em}
Note that this protocol and prompt design are directly taken from PhysGen3D. 
\vspace{1.0em}



\section{Analysis on SDE Hyperparameters}
\label{sec:appendix-hyperparam}

\paragraph{Regularization Parameter $\beta$.}
The regularization parameter $\beta$ balances the consistency term and the KL divergence term in Eq.~\ref{eq:objective}. As $\beta \to \infty$, the KL divergence term vanishes and the optimal distribution $p^{*}$ fails to preserve the original $q = \mathcal{N}((1-\tau)\boldsymbol{\mu}, {\tau}^2 \mathbf{I})$ (rightmost column of Fig.~\ref{fig:beta_ablation}). Conversely, as $\beta \to 0$, the consistency term vanishes and no consistency optimization occurs, producing non-photorealistic videos (leftmost column of Fig.~\ref{fig:beta_ablation}). A non-extreme $\beta$ is therefore required, and our choice $\beta = \frac{1-\tau}{\tau}$ is the most practical, as it removes the need for an explicit $v_{\epsilon}$.
\begin{figure}
    \centering
    \includegraphics[width=0.9\linewidth]{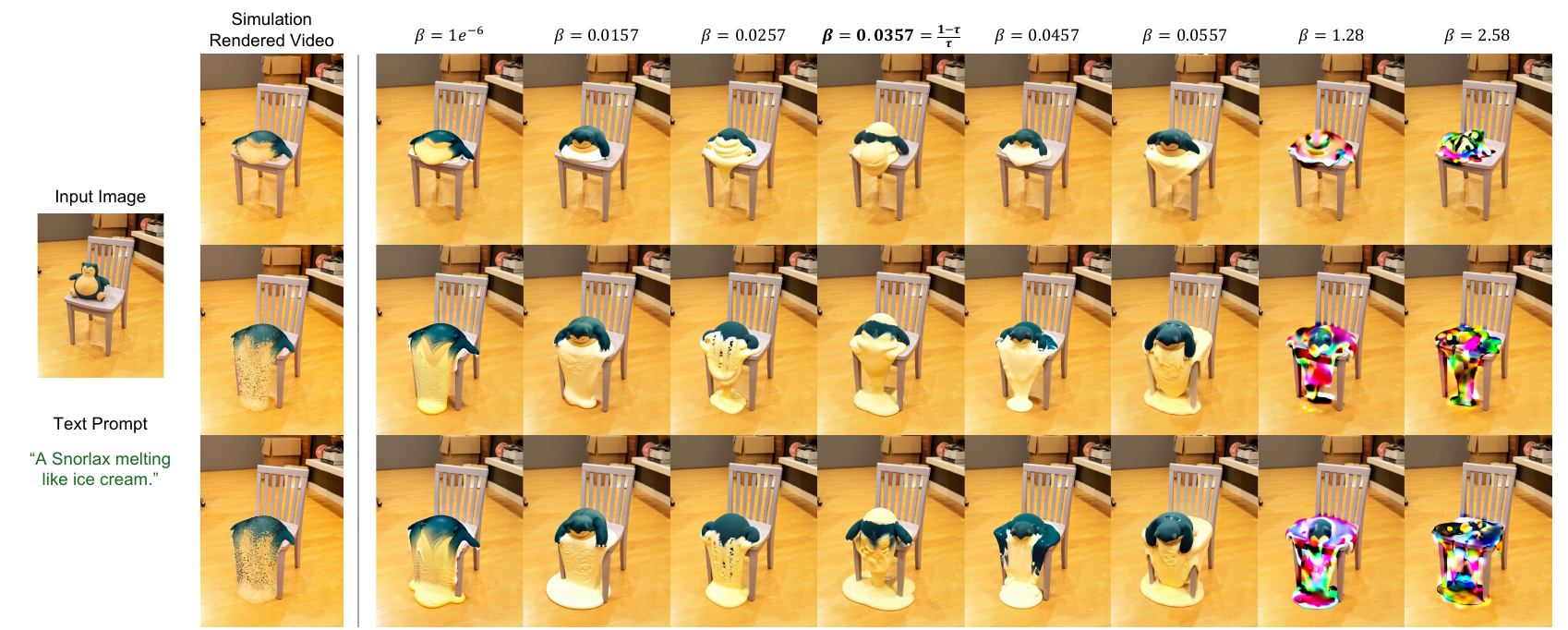}
    
    \caption{Comparison of generated videos across various $\beta$ values. $\beta=0.0357=\frac{1-\tau}{\tau}$ is our choice that entirely eliminates the $v_\epsilon$ term. $\beta=1e^{-6}$ represents the $\beta \to 0$ case, while $\beta=1.28$ and $\beta= 2.58$ represent the $\beta \to \infty$ case.}
    \label{fig:beta_ablation}

\end{figure}

\paragraph{Iterations $N$.}
The parameter specifies the number of SDE optimization steps applied via Eq.~\ref{eq:cf_formula} to achieve consistency with the input image. As shown in Fig.~\ref{fig:ablation}, increasing iterations progressively generates videos more consistent with the input image.
However, more iterations increase computational cost due to higher NFEs (Number of Function Evaluations) from repeated $v_\theta$ predictions. Empirically, we found that once the video model's prior reaches a satisfactory level of consistency, subsequent iterations maintain similar consistency without significant improvement. Therefore, we practically choose $N = 10$ to balance consistency requirements with computational efficiency, in conjunction with other hyperparameter settings for $\gamma$ and $\tau$.

\vspace{1.0em}

\begin{minipage}[t]{0.48\textwidth}
\strut
\textbf{Step Size $\gamma$.}
To ensure convergence of the Euler-Maruyama discretization~\citep{kloeden1977numerical, kloeden1992numerical}, the step size $\gamma$ should be sufficiently small. We conducted sensitivity experiments to verify the range of $\gamma$ values that guarantee this convergence property.
We evaluated various $\gamma$ values by monitoring whether the distribution of the optimized video latent during SDE optimization stably maintains the original distribution through the KL divergence term in our objective (Eq.~\ref{eq:objective}). Specifically, we tracked the norm of the latent at each SDE step, with results shown in Fig.~\ref{fig:step_size_sensitivity}.
Our experiments reveal that as SDE steps progress, the original distribution is well-preserved for $\gamma \leq 0.8$, whereas larger values fail to maintain distributional stability.
\end{minipage}%
\hfill
\begin{minipage}[t]{0.48\textwidth}
    \vspace{0pt}
    \strut
    \centering
    \includegraphics[width=1.0\linewidth]{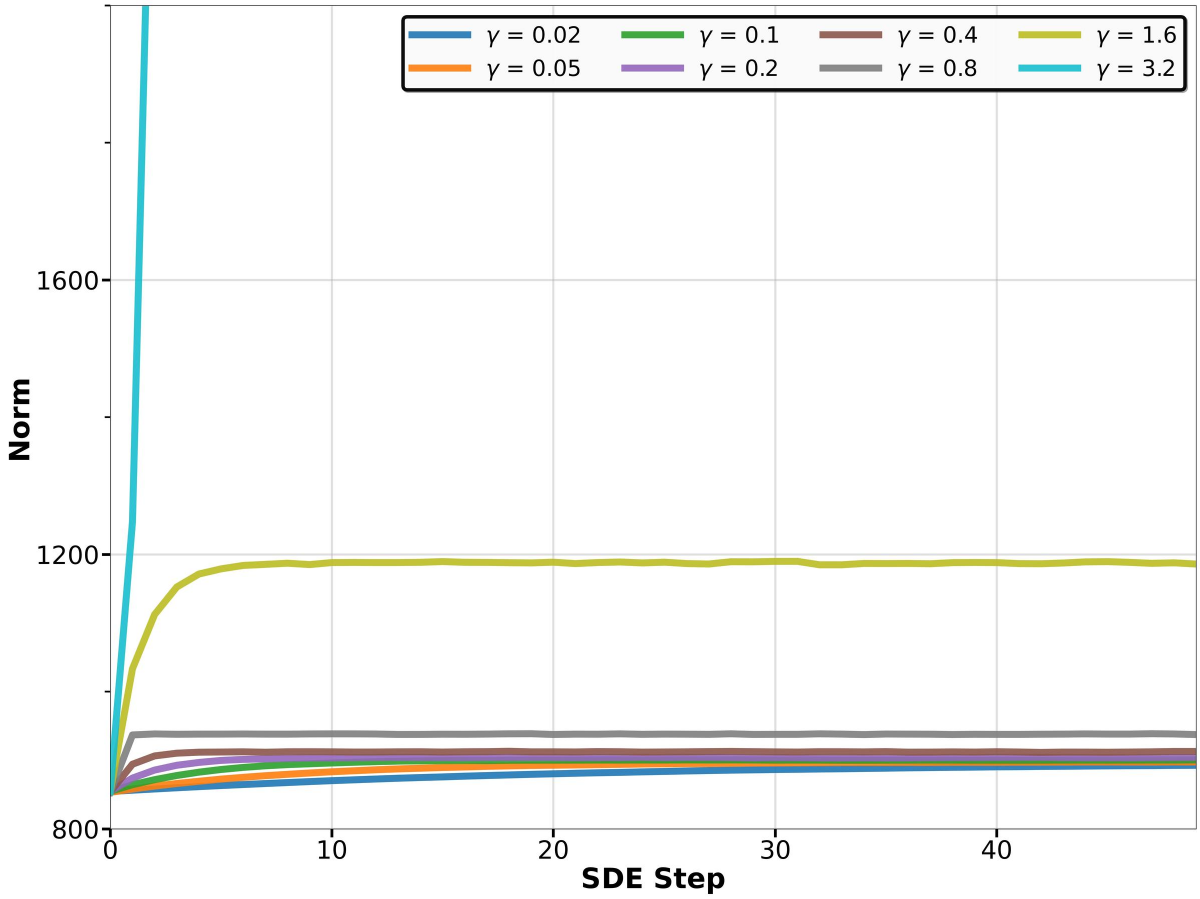}

    \captionof{figure}{Latent norm at each SDE optimization step for various step sizes $\gamma$.}
    \label{fig:step_size_sensitivity}
    \end{minipage}

\begin{figure}[h]
    \centering
    \includegraphics[width=0.9\linewidth]{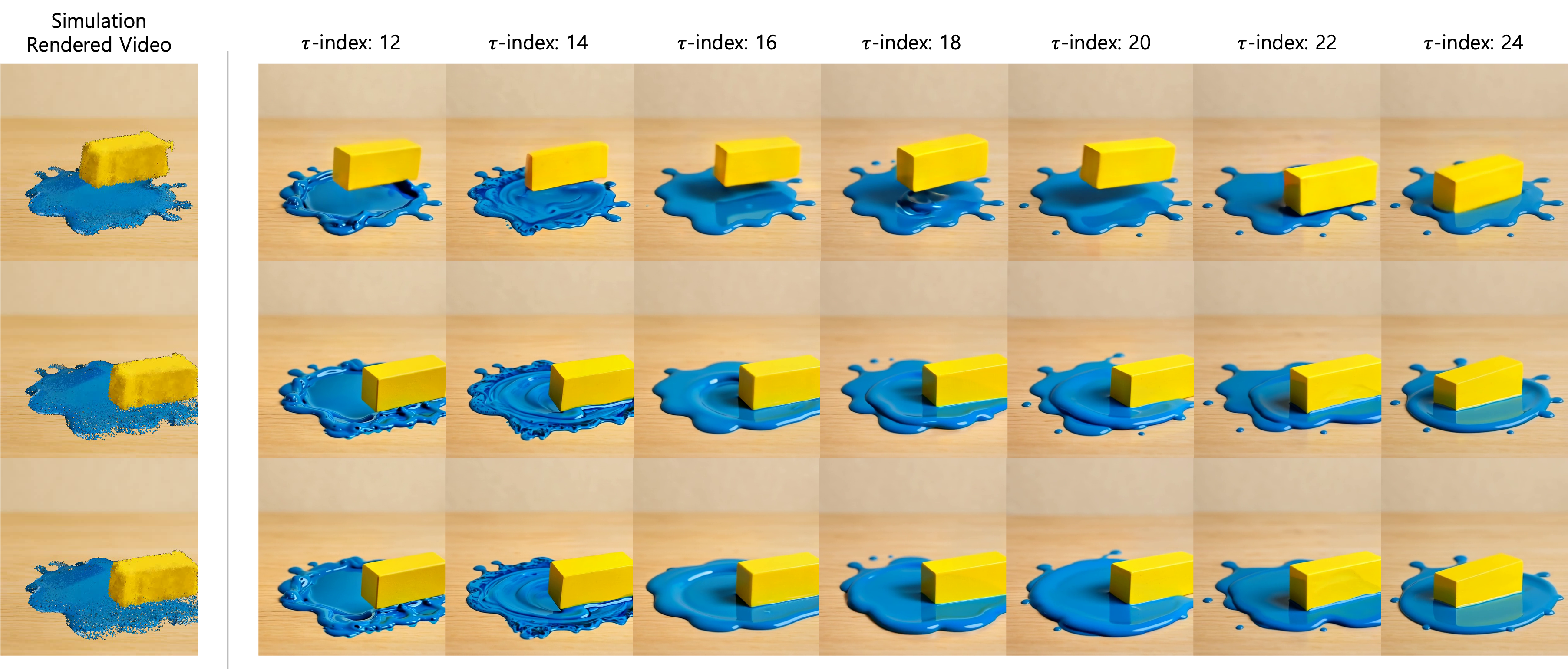}
    
    \caption{Comparison of generated videos across various target timesteps $\tau$. The $\tau$-index denotes the diffusion step corresponding to $\tau$ (total of 25 steps).}
    \label{fig:tau_ablation}

\end{figure}    

\paragraph{Target Timestep $\tau$.}
In the generation process of diffusion or flow models, early timesteps ($\tau \approx T$) focus on semantic structure (low-frequency components) while later timesteps ($\tau \approx 0$) refine details (high-frequency components). The choice of $\tau$ in our SDE optimization controls which aspects are optimized. Fig.~\ref{fig:tau_ablation} shows results across various $\tau$ values. At $\tau$-index $\approx 25$, the model relies too heavily on its semantic priors and ignores our guidance. At small $\tau$-index, the model cannot generate new components, failing to transform simulation-rendered point clouds into realistic liquids. At $\tau = 20$, the model balances guidance adherence with photorealistic rendering including shadows, liquid reflections, and surface details.

\section{Inference-Time Analysis}
\label{sec:appendix-inference}
In Tab.~\ref{tab:runtime}, we report an inference time breakdown of our pipeline, where each chunk produces 36 frames. All measurements are conducted on a single NVIDIA RTX~3090 GPU. Note that as the pipeline is fully training-free, it naturally benefits from high-quality, efficient off-the-shelf video models.

\paragraph{Stage~1 (overhead).}
Stage~1 performs mesh orbit rendering, including segmentation, point-cloud unprojection, mesh reconstruction, and rendering, which takes $\sim$\textbf{3.0 min}.

\paragraph{Stage~2 (overhead).}
Stage~2 performs physics simulation and point-based rendering with 50K particles, which takes $\sim$\textbf{1.3 min}.

\paragraph{Consistency-Guided Flow SDE.}
$\Phi_{\mathrm{CF}}$ requires 50 NFEs (20 for inversion, 10 for SDE optimization, and 20 for final generation). With a measured cost of 5.1629\,s per NFE, it takes $\sim$\textbf{4.3 min}.
\vspace{1.0em}

\begin{table}[h]
\centering
\footnotesize
\setlength{\tabcolsep}{10pt}
\renewcommand{\arraystretch}{1.2}
\begin{tabular}{l|cc|c}
\toprule
\textbf{Component} & \textbf{Stage~1 (overhead + $\Phi_{\mathrm{CF}}$)} & \textbf{Stage~2 (overhead + $\Phi_{\mathrm{CF}}$)} & \textbf{Total (1st Chunk)} \\
\midrule
\textbf{Time} & $\sim$7.3 min & $\sim$5.6 min & $\sim$12.9 min \\
\bottomrule
\end{tabular}
\vspace{1.5em}
\caption{Runtime breakdown.}
\label{tab:runtime}
\end{table}


\section{Limitations \& Future Work}
\label{app:failure}

\textsc{3DPhysVideo} achieves strong results across diverse physical scenarios, while several limitations remain.
\textbf{(i)} Errors in single-view 3D reconstruction using VGGT propagate to the simulation.
\textbf{(ii)} Errors in VLM-based physical parameter estimation yields unrealistic videos (see Fig.~\ref{fig:rebuttal_failure}).
Furthermore, like other explicit simulator-based methods~\citep{liu2024physgen, tan2024physmotion, chen2025physgen3d, li2025wonderplay,physctrl2025, zhan2026perpetualwonder, foo2026physical}, our
framework is limited in that
\textbf{(iii)} objects that newly appear in subsequent frames are
not handled, and
\textbf{(iv)} self-propelled objects (e.g., living agents) lie
outside the simulator's scope.
Finally, although efficient on a consumer GPU
as shown in Tabs.~\ref{table:stage1},~\ref{table:stage2}, and~\ref{tab:runtime},
\textsc{3DPhysVideo} is \textbf{(v)} not yet real-time.

Addressing these limitations requires more accurate 3D prior models, generic physics simulators integrated with AI agents, and few-step distillation for real-time inference, which we leave as future work.

\begin{figure}[h]
    \centering
    \includegraphics[width=1.0\linewidth]{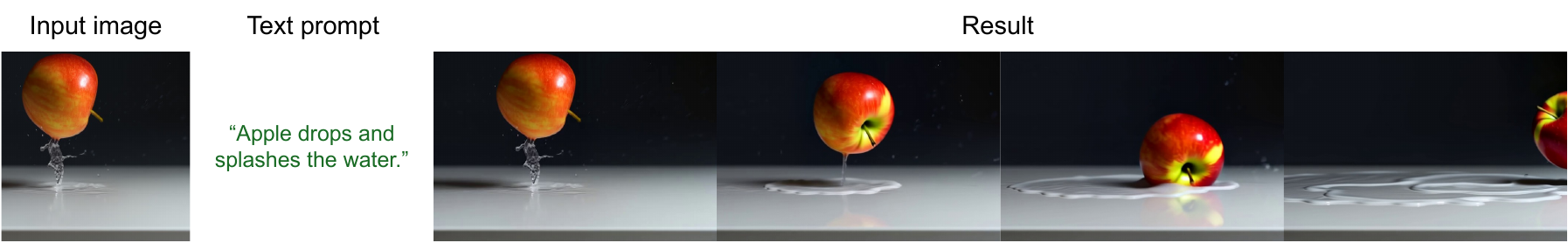}
    \caption{Failure case. An apple with an excessively low Young's modulus ($1\times10^{6}$~Pa) estimated by the VLM deforms unrealistically. Upon hitting the surface, it squashes like a soft, jelly-like object rather than a firm fruit.}
    
    \label{fig:rebuttal_failure}
\end{figure}

\paragraph{Broader Impacts.}
\label{app:broader_impacts}
\textsc{3DPhysVideo} can support positive applications in scientific visualization, education, virtual content creation, and physics-based animation by lowering the bar to producing physically plausible video content from a single image. As with all video generation methods, however, our work also carries potential risks of misuse, such as the creation of deceptive or misleading visual content. Note the proposed method remains compatible with watermarking and provenance-tracking mechanisms developed for generative videos.

\section{More Results from \textsc{3DPhysVideo}}
\label{app:more-results}

\begin{figure}[h]
    \centering
    \includegraphics[width=1.0\linewidth]{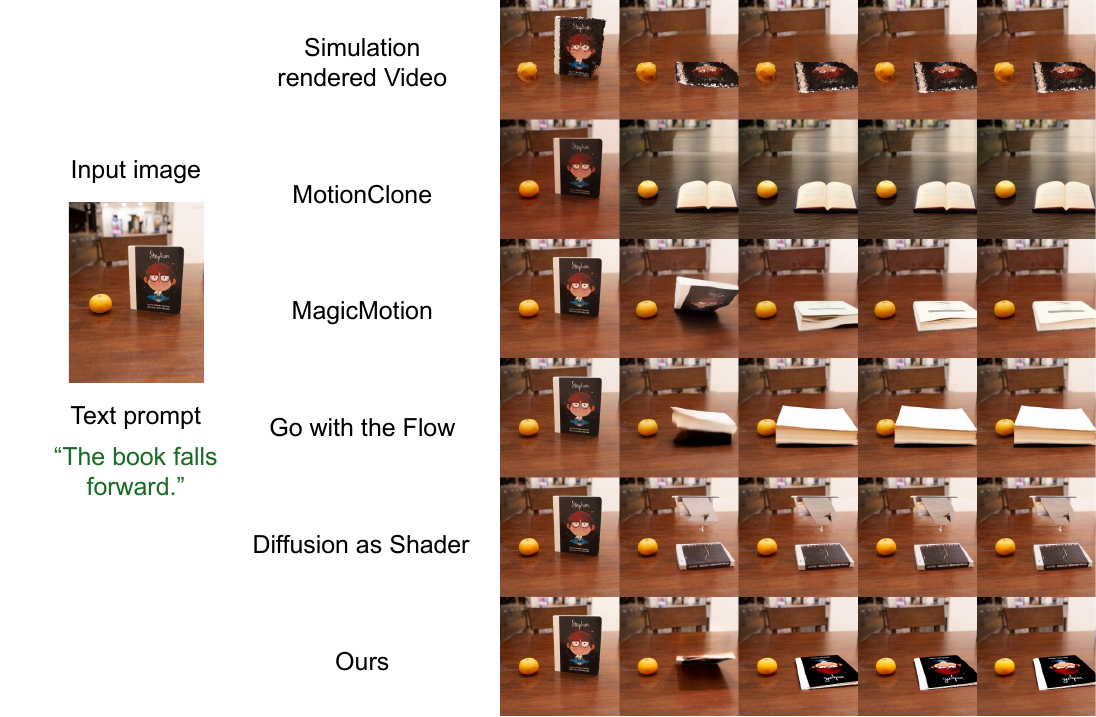}
    \caption{Qualitative comparison with state-of-the-art simulation to video models on the book falling scene. }
    \label{fig:placeholder}
\end{figure}

\begin{figure}[h]
    \centering
    \includegraphics[width=1.0\linewidth]{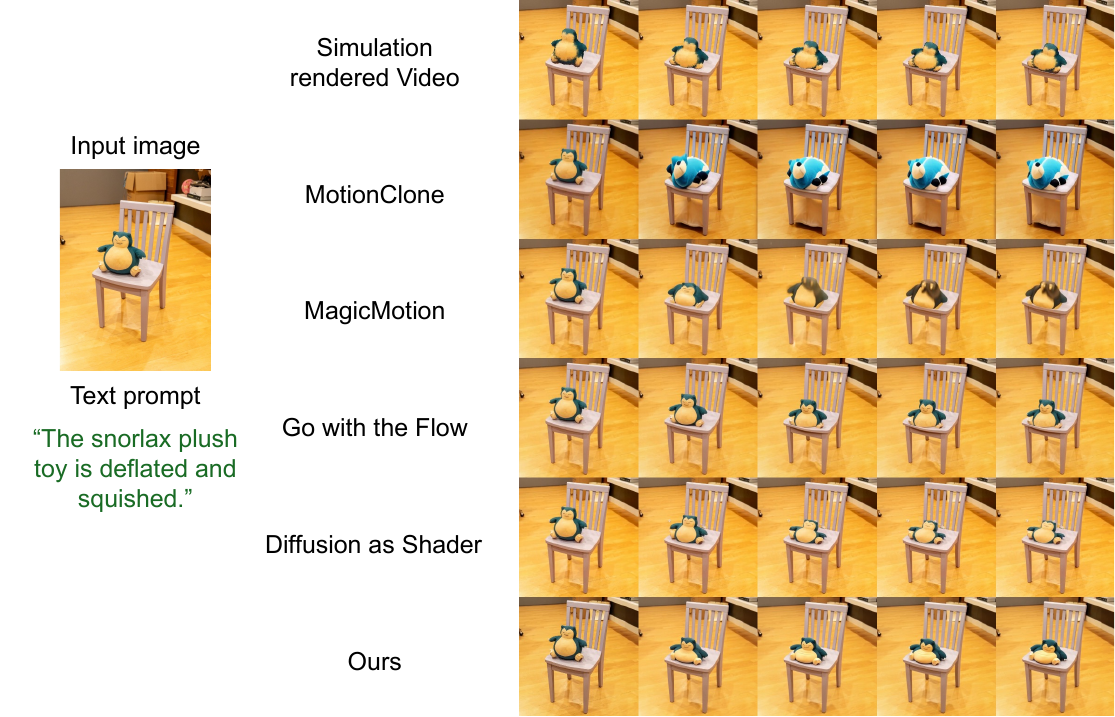}
    \caption{Qualitative comparison with state-of-the-art simulation to video models on the Snorlax deflating scene. }
    \label{fig:placeholder}
\end{figure}

\begin{figure}
    \centering
    \includegraphics[width=1.0\linewidth]{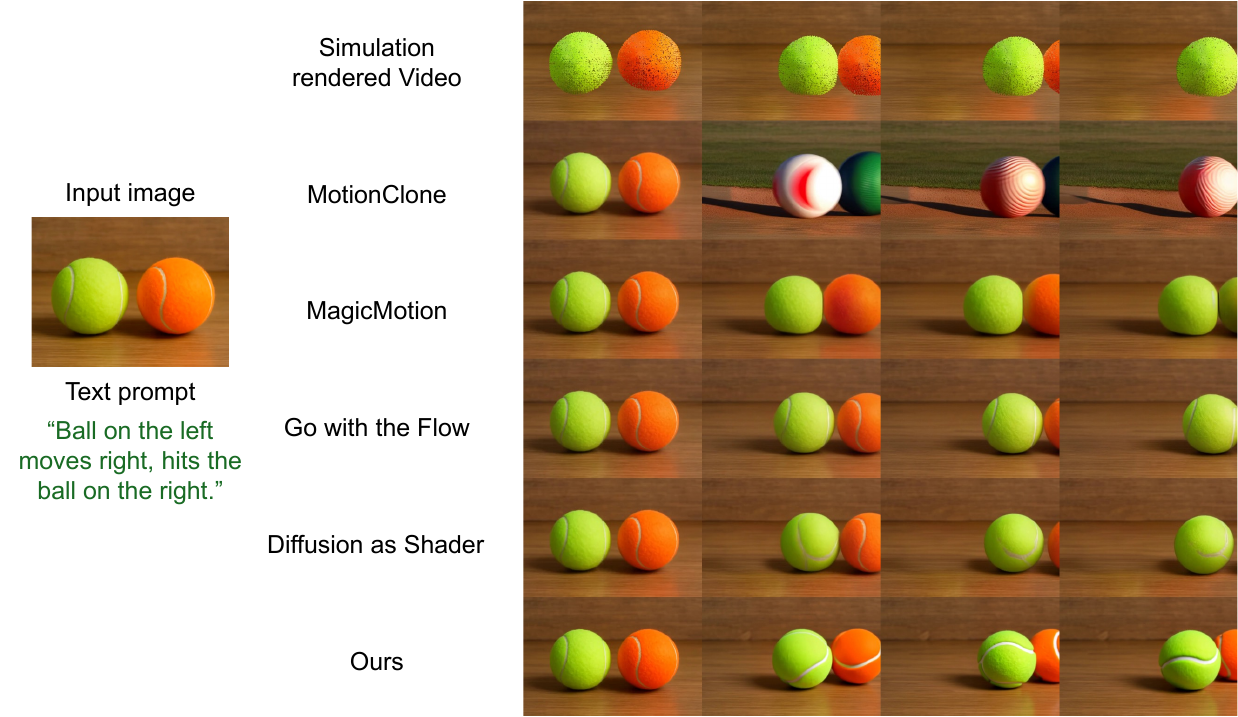}
    \caption{Qualitative comparison with state-of-the-art simulation to video models on the ball collision scene. }
    \label{fig:placeholder}
\end{figure}

\begin{figure}
    \centering
    \includegraphics[width=1.0\linewidth]{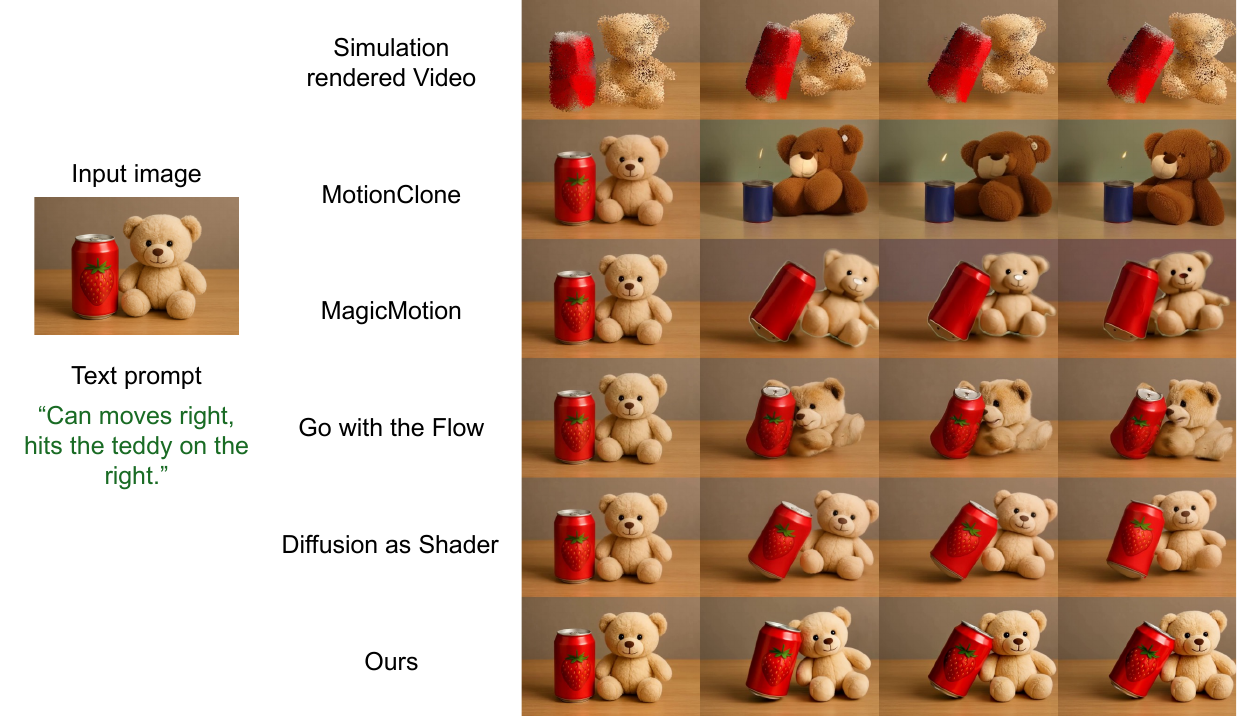}
    \caption{Qualitative comparison with state-of-the-art simulation to video models on the can-teddy collision scene.}
    \label{fig:placeholder}
\end{figure}


\begin{figure}
    \centering
    \includegraphics[width=1.0\linewidth]{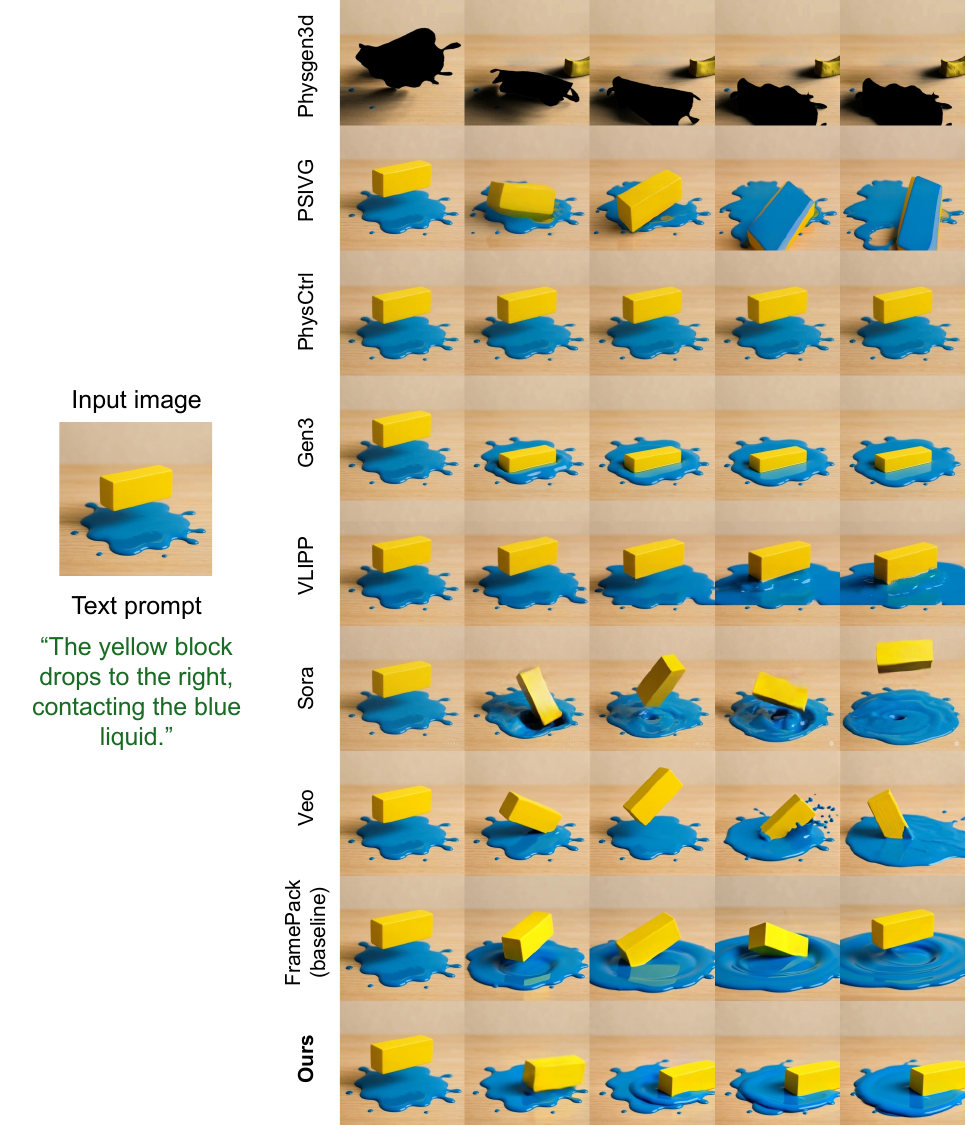}
    \caption{Qualitative comparison on the block falling scene.}
    \label{fig:placeholder}
\end{figure}

\begin{figure}
    \centering
    \includegraphics[width=1.0\linewidth]{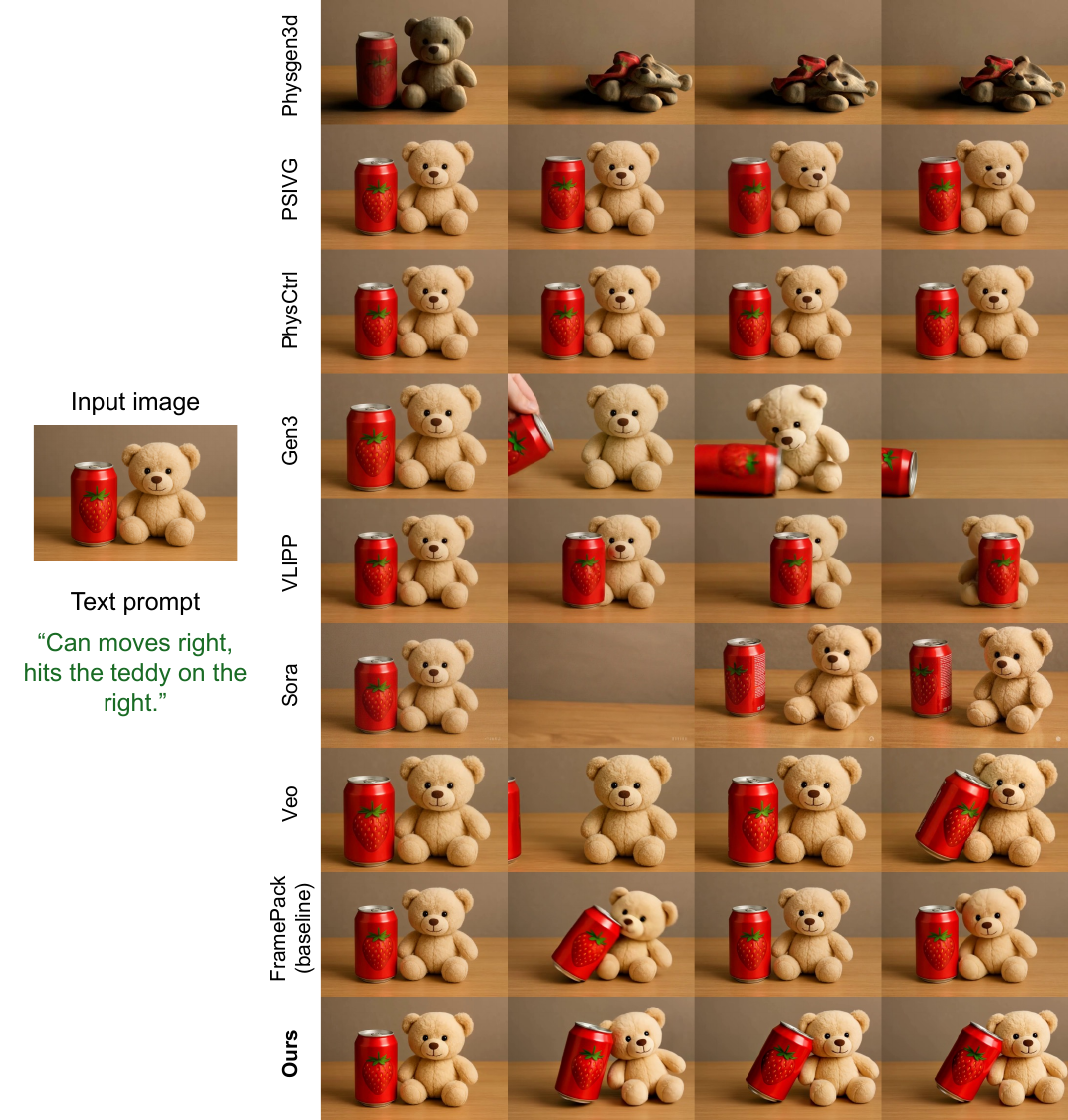}
    \caption{Qualitative comparison on the synthesized can-doll collision scene.}
    \label{fig:placeholder}
\end{figure}

\begin{figure}
    \centering
    \includegraphics[width=1.0\linewidth]{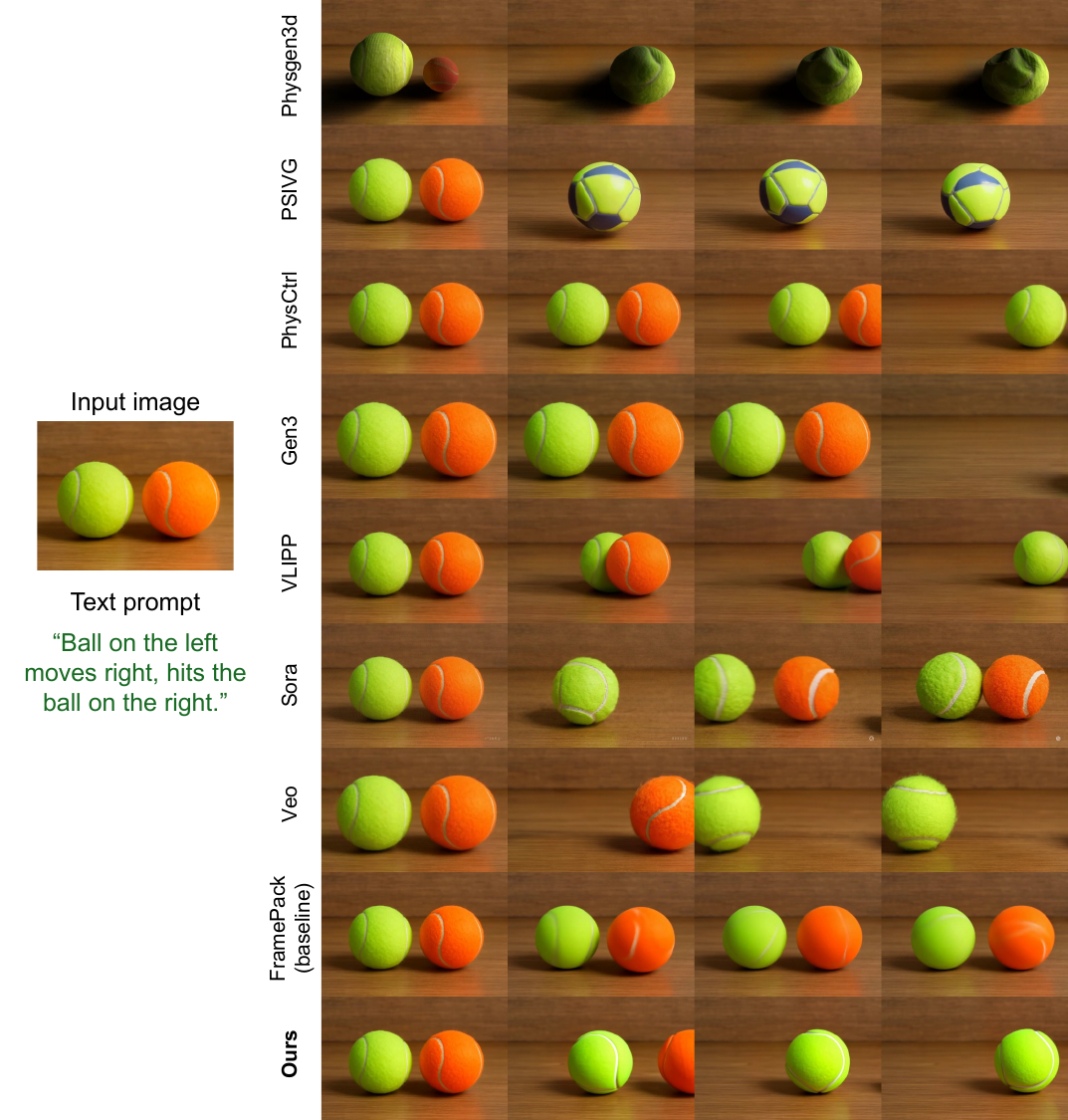}
    \caption{Qualitative comparison on the ball collision scene.}
    \label{fig:placeholder}
\end{figure}

\begin{figure}
    \centering
    \includegraphics[width=1.0\linewidth]{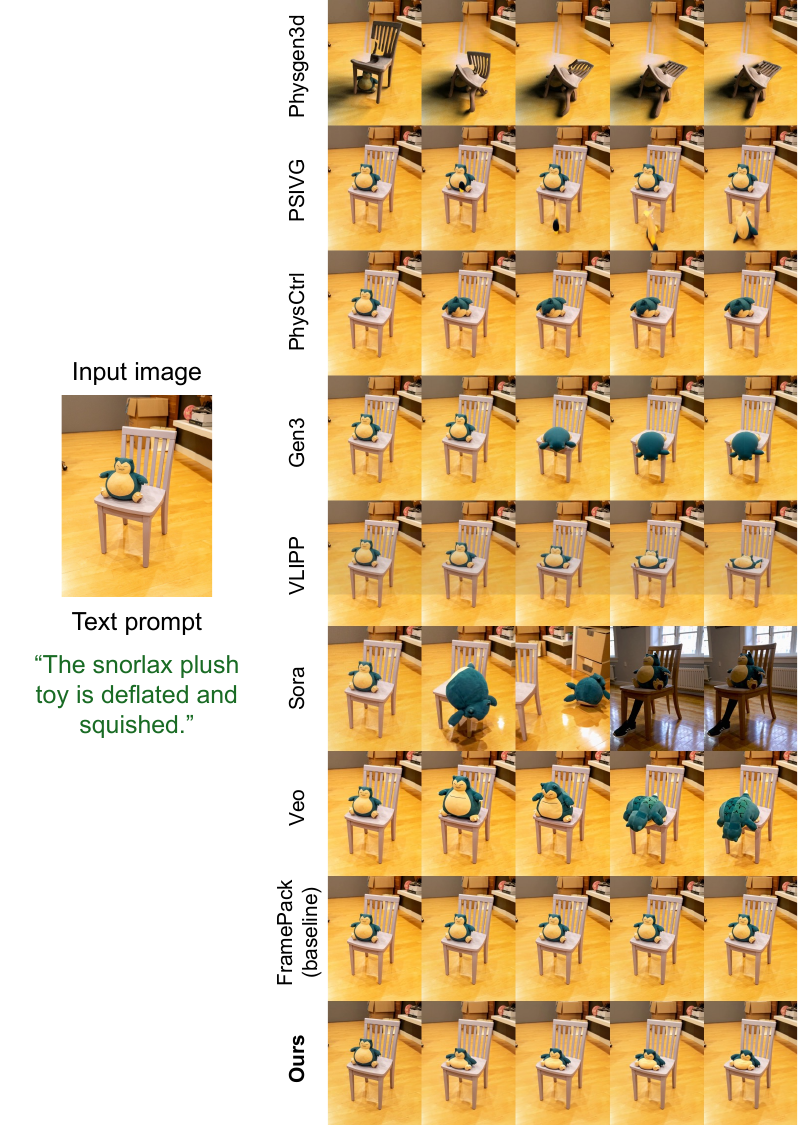}
    \caption{Qualitative comparison on the snorlax deflating scene.}
    \label{fig:placeholder}
\end{figure}


\FloatBarrier  
\clearpage

\end{document}